\documentclass[accepted,3p]{elsarticle}

\journal{MSSP}

\usepackage{url}
\usepackage{amsmath}
\usepackage{amssymb}

\usepackage[binary-units=true]{siunitx}
\usepackage[nolist]{acronym}
\usepackage{algorithm}
\usepackage[noend]{algpseudocode}

\usepackage{multirow}
\usepackage{makecell}
\usepackage{pgfplotstable}
\usepackage{array}
\usepackage{booktabs} 
\usepackage{siunitx} 

\usepackage{nicefrac}

\usepackage{hyperref}
\usepackage{cleveref}

\usepackage{subcaption}
\usepackage{graphicx}
\graphicspath{{fig/}}

\newcommand*\mean[1]{\bar{#1}}
\def\BState{\State\hskip-\ALG@thistlm}

\usepackage{changes}
\newcommand{\changed}[1]{{#1}}
\PassOptionsToPackage{normalem}{ulem}
\usepackage{ulem}
\newcommand{\removed}[1]{}

\begin{document}
	\begin{frontmatter}
		\title{A general anomaly detection framework for fleet-based condition monitoring of machines}

		\author[SISW,CW]{Kilian Hendrickx\corref{cor1}}
		\ead{kilian.hendrickx@siemens.com}
		\author[CW]{Wannes Meert}
		\author[SISW,ULB]{Yves Mollet}
		\author[ULB]{Johan Gyselinck}
		\author[SISW]{Bram Cornelis}
		\author[PMA,FM]{Konstantinos Gryllias}
		\author[CW]{Jesse Davis}  
	
		\address[SISW]{Siemens Digital Industries Software, Interleuvenlaan 68, 3001 Leuven, Belgium}
		\address[CW]{KU Leuven, Department of Computer Science, Celestijnenlaan 200A box 2402, 3001 Leuven, Belgium}
		\address[ULB]{Universit\'{e} Libre de Bruxelles, BEAMS, Avenue Franklin Roosevelt 50 (CP165/52), 1050 Brussels, Belgium}
		\address[PMA]{KU Leuven, Department of Mechanical Engineering, Celestijnenlaan 300, 3001 Leuven, Belgium}
		\address[FM]{Dynamics of Mechanical and Mechatronic Systems, Flanders Make, Belgium}
		
		\begin{abstract}
			Machine failures decrease up-time and can lead to extra repair costs or even to human casualties and environmental pollution. Recent condition monitoring techniques use \changed{artificial intelligence} in an effort to avoid time-consuming manual analysis and handcrafted feature extraction. Many of these only analyze a single machine and require a large historical data set. In practice, this can be difficult and expensive to collect. However, some industrial condition monitoring applications involve a fleet of similar operating machines. In most of these applications, it is safe to assume healthy conditions for the majority of machines. Deviating machine behavior is then an indicator for a machine fault.
			
			This work proposes an unsupervised, generic, anomaly detection framework for fleet-based condition monitoring. It uses generic building blocks and offers three key advantages. First, a historical data set is not required due to online fleet-based comparisons. Second, it allows incorporating domain expertise by user-defined comparison measures. Finally, contrary to most black-box \changed{artificial intelligence} techniques, easy interpretability allows a domain expert to validate the predictions made by the framework. 
			
			Two use-cases on an electrical machine fleet demonstrate the applicability of the framework to detect a voltage unbalance by means of electrical and vibration signatures.
		\end{abstract}
		
		
	\end{frontmatter}
	
	\begin{acronym}
		\acro{CM}{Condition Monitoring}
		\acro{DTW}{Dynamic Time Warping}
		\acro{MCSA}{Motor Current Signature Analysis}
		\acro{FFT}{Fast Fourier Transform}
		\acro{SCIM}{Squirrel cage induction motor}
		\acro{WRSM}{Wound Rotor Synchronous Motor}
		\acro{FFT}{Fast Fourier Transform}
		\acro{DC}{Direct Current}
		\acro{kNN}{k-th Nearest Neighbor}
		\acro{LOF}{Local Outlier Factor}
		\acro{ANN}{Artificial Neural Networks}
		\acro{DL}{Deep Learning}
		\acro{SVM}{Support Vector Machine}
	\end{acronym}
	
	\section{Copyright information}
	Accepted manuscript for Mechanical Systems and Signal Processing (\url{https://www.journals.elsevier.com/mechanical-systems-and-signal-processing/})
	
	\textcopyright 2019. Licensed under the Creative Commons CC-BY-NC-ND (\url{https://creativecommons.org/licenses/by-nc-nd/4.0/})
	
	\section*{Highlights}
	\begin{enumerate}
		\item A framework is proposed for fleet-based condition monitoring
		\item Online fleet-based comparisons avoid the need for a historical data set
		\item It is easy for a domain expert to interact with this framework 
		\item A fleet of electrical drivetrains is used to experimentally validate the framework
		\item The framework is benchmarked against a classic signal processing approach
	\end{enumerate}
	
	\section{Introduction}
		\subsection{Context}
			\label{sec:introduction_context}
			The performance and reliability of machines are important for several industries, as in certain circumstances failures or malfunctions may lead to high production losses, severe injuries or even loss of lives. Condition monitoring allows for the early detection of machine failures, which helps to avoid further damage and/or to reduce downtime. Many condition monitoring approaches are limited to analyzing a single machine \cite{Wong2006}. However, some real-world industrial applications could benefit from monitoring multiple similar machines or entities (i.e., a fleet) simultaneously \cite{Lee2014,Matthews2013}. This need arises in wind turbines \cite{Siegel2013}, production lines, \changed{and} aerospace engines (which often operate in pairs) \cite{Jacobs2018}. \changed{Moreover, in fleet settings like monitoring the components of a train or a conveyor belt (e.g. the traction motors, bearings and bogies \cite{Hodge2015,Fumeo2015}), the machines in the fleet are operating under similar conditions.}
			A fleet-based approach leverages the fact that multiple entities are analyzed. In each of the previous examples, the different entities have comparable signature characteristics which deviate when a fault appears. A fault indicator can use this deviation.
		
		\subsection{Need for \changed{AI-based} fleet monitoring approaches}
			\label{sec:introduction_need}
			The traditional condition monitoring approach involves handcrafting an indicator of the phenomena of interest, which requires significant expertise in machine signature analysis and machine usage. In deployment, the current value of the indicator is compared to a pre-selected threshold to decide on the health condition of a machine. In practice, this approach has several drawbacks. First, operational conditions can influence machine behavior, which requires either setting different thresholds \cite{Siegel2013,Meinguet2014}, identifying dedicated operational conditions for fault detection (e.g. ISO 10816-3:2009 \cite{ISO108163200}) or normalizing the fault indicator \cite{Farrar2001}. This approach is not scalable to a large number of operational conditions. Second, a machine can be sensitive to multiple fault types. Thus, a robust condition monitoring approach requires a manual tuning procedure for each expected machine fault and their combinations. For complex systems, it is often not possible to know all these upfront. For this reason, an alternative approach compares machine data with a simulation model of healthy behavior \cite{Jia2018}. However, this assumes that every considered input parameter and the machine behavior does not vary over time. 
			
			Recent condition monitoring studies often suggest \changed{employing techniques from artificial intelligence as an alternative to these handcrafted solutions. In particular, the two most commonly mentioned subfields of AI are machine learning and data mining~\cite{Siegel2013,Lee2014a,Liu2018,Zhao2019}. Typically, researchers consider supervised, semi-supervised and unsupervised techniques from these two subfields.} 
				
			\changed{Supervised approaches to fault detection such as \ac{ANN} \cite{Yang2004,Unal2014,Schlechtingen2011}, \ac{DL}			\cite{Zhao2019,Jia2016,Khan2018,Chen2019} and \ac{SVM} \cite{Ge2004,Widodo2007,Gryllias2012} require large historical labeled data sets. That is, the data must be annotated with the true machine health condition. Moreover, they often assume the data contains examples of every possible fault type \cite{Mauricio2017}. However, such data sets are not always available in industrial environments \cite{Bull2018,Rogers2019}. In those applications, even a limited historical data set such as the \cite{Zhang2013} proposes can be challenging to obtain.} As an alternative, a physical model can simulate faults \cite{Gryllias2012,Sobie2018}, but this is often very challenging in practice as not every possible fault can be accurately modeled.
			
			To alleviate the need to collect data about faults, another approach is to consider semi-supervised techniques, like novelty detection \cite{Pimentel2014,Gryllias2015}. These techniques simply \changed{use a historical data set containing only healthy machines} build a model of healthy behavior. When deployed, they detect behaviors that deviate compared to the model of normal behavior \changed{\cite{Abu-El-Zeet2006, West2012,Li2015}.} Consequently, it is very important that data from a faulty machine is not present in the training data. Unfortunately, this is hard to guarantee in reality as machines degrade over time. Moreover, this assumes that normal behavior does not change over time, which may not be the case. \changed{An alternative approach is to cluster historical data and assign a label to each cluster. This allows classifying new machine measurements by assigning it the label of its nearest cluster. However, this approach requires expert knowledge to provide the labels \cite{Al-Dahidi2018a}.}
						
			Finally, unsupervised anomaly detection can address the drawbacks of the supervised and semi-supervised approaches \cite{Chandola2009,Pontoppidan2005,Purarjomandlangrudi2014,Vercruyssen2018}. These techniques require neither labeled data nor solely healthy behavior. Instead, unsupervised techniques assume that the majority of the data set belongs to the normal (healthy) class and identifies rare or infrequent behaviors as abnormal (faulty). Still, using these techniques poses two challenges in practice.
			First, they require selecting a threshold to determine the boundary between normal and anomalous behavior, which is challenging as these behaviors may overlap due to noise and uncertainties. Setting this parameter requires reasoning about the cost of misclassifying a faulty machine as healthy (and vice versa) and should thus be done by a domain expert. 
			Second, in a condition monitoring context, it is often necessary to collect an anomaly detection data set of machines in similar operational conditions, especially when a historical data set is not available.
			
			\changed{A drawback shared by many of the existing approaches is that they employ black-box models, and hence it is difficult to gain insight into why a model made a certain prediction. Consequently, domain experts may not trust the predictions that the model is making. Moreover, it is non-trivial to incorporate expert knowledge into models such as \ac{ANN}s.}
			
		\subsection{\changed{Our Contribution}}
			\label{sec:introduction_task}
			\changed{
				This paper's contribution is an artificial intelligence based general condition monitoring approach. The framework simultaneously analyzes a fleet of machines under the assumption that each machine is in a comparable operational condition. To perform the analysis, it uses three inter-related generic blocks: comparing pairs of machines, clustering the entire fleet of machines, and identifying anomalous machines. Each block should have a meaningful outcome that can be visualized to give a domain expert an overview related to the observed machine behavior.
				First, we discuss several different ways to compare pairs of machines. Most interestingly, we consider a distance measure based on the amount of warping that occurs in a dynamic time warping. While measures based on Dynamic Time Warping are common in the AI literature, these receive less attention in the context of mechanics and condition monitoring.
				Second, we employ clustering which is a core artificial intelligence technique~\cite{Russell2009}. While the clustering can be done with a variety of algorithms, we recommend using a hierarchical method because of its simplicity and interpretability. We then propose an approach to partition a hierarchical clustering into a flat one that is novel within the context of machinery diagnostics.
				Finally, to identify potentially anomalous machines, we exploit the reasonable assumption that the majority of the machines in the fleet are healthy. Consequently, a good indication of a fault is when a machine's current behavior deviates from the behavior currently exhibited by the majority of the machines. Like all condition monitoring and anomaly detection applications, this assumes that a fault alters the data generated by a machine~\cite{Aggarwal2017}.
				Empirically, we consider machinery diagnostics tasks for a fleet of electrical drive trains. We examine two use-cases (electrical and vibration signatures), each with several variants that use differing amounts of domain knowledge. Applying fleet-based condition monitoring to electrical signature analysis is novel, particularly in terms of the fact that we analyze the raw time-series data. We find that our AI-based framework outperforms the more tradition condition monitoring approach the vast majority of the time.

				The proposed methodology offers four advantages over existing approaches.
				First, unlike other approaches, the framework does not require any historical data. Instead, it makes use of online comparisons among machines. However, if they are available, recent machine measurements can improve performance by extending these comparisons.
				Second, the framework enables interpretable visualizations of the inter-related generic blocks. These allow a domain expert to gain insight into the framework's predictions. Hence, this work considers relatively simple but intuitive techniques.
				Third, exploiting relevant domain knowledge allows tailoring the framework towards the specific use-case. For example, a measure to compare a machine pair's behavior can be based on mechanical knowledge. This is not possible with techniques such as neural networks and deep learning.
				Finally, the framework can be applied directly on time-series data (i.e. raw measurements) and does not require a feature-based representation. However, the framework can use these features if available.
			}

			The rest of the paper is organized as follows. \Cref{sec:framework} proposes the fleet-based condition monitoring framework. Multiple possible implementations are discussed for each of the building blocks. Next, the framework is applied and evaluated on two use-cases considering a fleet of electrical drive trains. \Cref{sec:use_case_vu_curr} processes the electrical signatures in three different signal domains, while \cref{sec:use_case_vu_acc} analyzes vibration data. Both experimental evaluations are compared to handcrafted signal processing techniques for condition monitoring as a benchmark. Finally, \cref{sec:conclusion} discusses some general insights and conclusions.
	
	\section{General fleet-based condition monitoring framework}
		\label{sec:framework}
		An anomaly detection framework for fleet condition monitoring of similar machines is proposed, which detects deviating, thus probably faulty, machine behavior. This approach can detect various faults, even those not considered a priori, and does not require a training data set. 
		The fault detection process relies on three key assumptions. First, the majority of the machines are healthy, machines whose behavior deviates from the majority are considered faulty. Second, all machines are operating in a similar operational condition, having \changed{an} identical machine signature (e.g., speed). Third, the machines are subject to similar relevant environmental parameters influencing the machines' behavior. If this were not the case, deviating behavior could arise due to a different environment instead of a machine's health status. \changed{Practical examples where all of these assumptions hold are applications such as monitoring a conveyor belt or railway components. In both of these cases, a deviating machine is an indication of a machine fault.
			
		Thoughtful preprocessing could allow utilizing the proposed framework in many more applications by considering external or contextual variables as well as domain knowledge.
		First, a heterogeneous fleet can cause machine diversity. However, expert knowledge such as an ontology could help to select comparable subsets of machines \cite{Medina-Oliva2012,Medina-Oliva2014,Voisin2011}. The framework could then assess each of these subsets individually.
		Second, the machine fleet can operate in dissimilar operational conditions. In such a case, two strategies could allow machine comparisons: normalization and a short-term historical database. The former exploits domain knowledge. For example, if the physics of a machine is available, then it is possible to normalize the measurements with respect to the working conditions such as by removing the effect of speed using angular resampling~ \cite{Farrar2001}. The latter contains recent machine measurements in various operational conditions. Clustering operational parameters could provide relevant data samples of machines in comparable operational conditions, which the framework would then use to assess their current health status'.
		Third, the machine fleet might be subject to different environmental parameters. Clustering machines based on these parameters could result in identifying subsets of comparable machines.}
		
\begin{figure}
			\centering
			\includegraphics[width=0.5\textwidth]{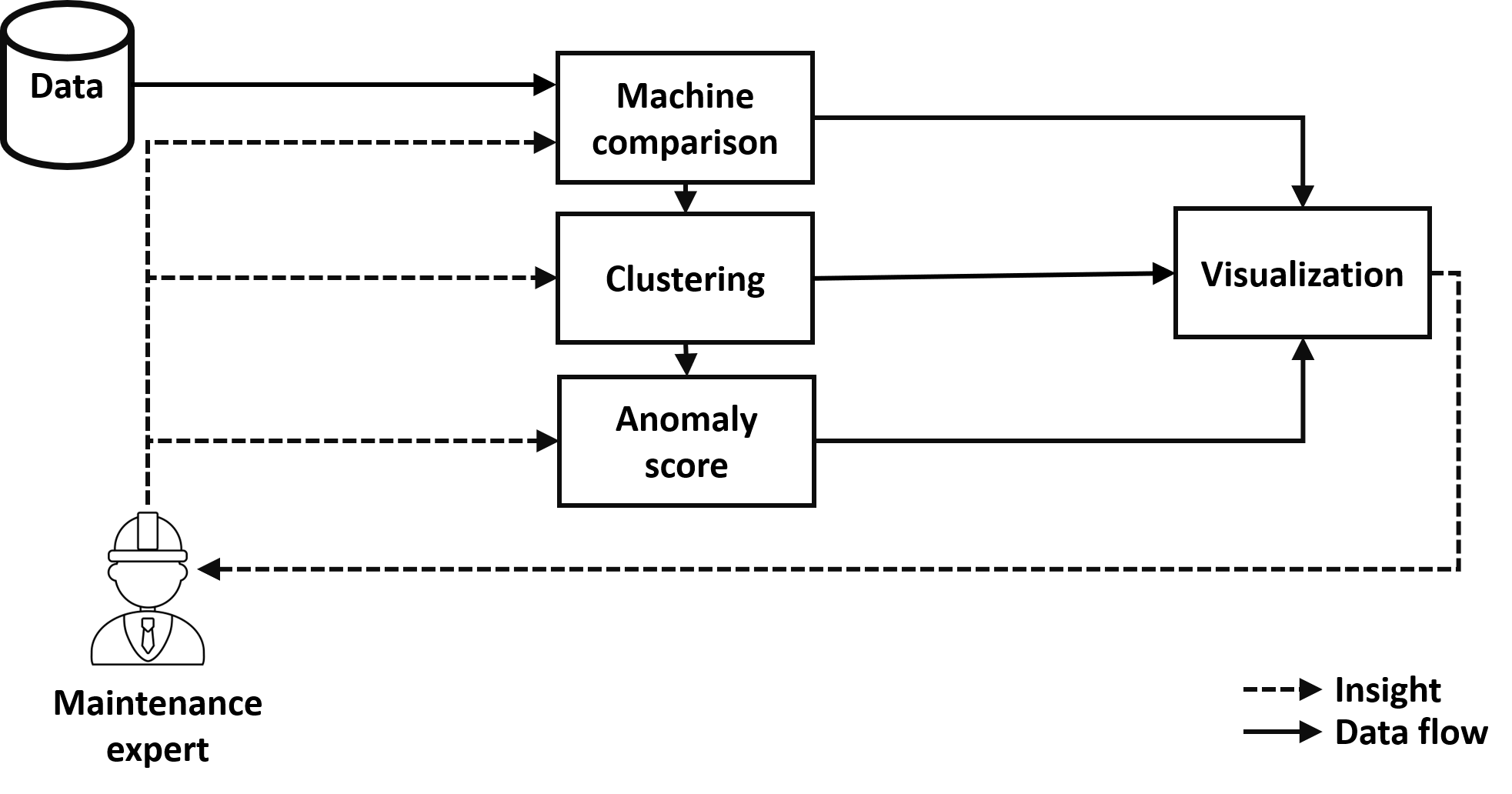}
			\caption{Interaction of the four blocks. Visualization can aid in interpreting the predictions. A user can select the machine comparison and anomaly score blocks based on domain knowledge, if available.}
			\label{fig:framework_blocks}
		\end{figure}

		Four building blocks interact in the proposed framework: machine comparison, fleet clustering, anomaly detection, and visualization (\cref{fig:framework_blocks}). The design and implementation choices of each block can be adapted to the application's need, based on expert knowledge.
		The first block measures the similarity between two machine behaviors, where a variety of measures are possible. While this work discusses several concrete possibilities, a user can select an appropriate one for the application at hand based on domain knowledge. 
		The second block uses a clustering algorithm with the selected measure to group\removed{ together} machines that are behaving similarly.
		The third block uses the discovered clusters to assign an anomaly score to each machine. In a fleet where the majority of the machines are healthy, a faulty machine could occur in any cluster that consists of less than half the machines. The fourth block provides a deeper understanding of the framework by visualizing the result of the other blocks to help guide a domain expert to analyze a specific deviating machine. 
		
		The framework provides \changed{an} online fleet analysis by considering small analysis windows. The required window size depends on the implementation of the first block, for example when evaluating in the frequency domain. The robustness of the prediction can be improved by considering a machine faulty based on $n$ consecutive analysis windows.
		
		\subsection{Building block 1: Comparing two machines}
			\label{sec:machine_comparion}
			Clustering uses the similarity of entities to infer a number of groups (i.e., clusters). Deciding how to compare a pair of machines $(X, Y)$ is thus an important step in the proposed framework. Optionally, it allows exploiting engineering knowledge to guide the clustering. 
			This block consists of two actions. First, the machine data is transformed into the desired signal domain. Second, the machines are compared in this new domain.
			Sequential data is represented as $(x_1, x_2, \ldots, x_n)$ where $x_i$ is one data point and $n$ is the length of the sequence. Each machine in the fleet is assumed to have the same total number of data points $n$. 
			
			\subsubsection{Domain specific preprocessing}
				Data about machines is generated by sensors that capture measurements in the time domain. Transforming the raw data to another signal domain can emphasize certain physical phenomena.
				The framework can support any type of transformation. This work focuses on four output domains: time, frequency, time/frequency, and a feature space.
				
				For all domains\changed{,} it is important to normalize the data before comparing machines to help remove differences that simply arise due to the unique characteristics of each machine and cope with the fact that the measurement values can originate from different mechanical units. Without normalization, a comparison between machines would be dominated by the unit with the largest range.
			
				\paragraph{Time domain}
					\label{par:block1_preprocessing_time_domain}
					Waveforms can be used to analyze time domain measurements, as they contain information about amplitude, shape, and existing symmetries. Time domain analysis is particularly interesting to detect trends and patterns over time.
					
					Time series normalization is used to scale all machine measurements to the same range. Methods such as min-max scaling, ${(X-min(X))}/{(max(X)-min(X))}$, and z-score scaling, ${(X - \mean{X})}/{X_\sigma}$, are popular, but sensitive to outliers and should be used with care. If outliers are expected, more robust estimators like percentile scaling, $({X - median(X))}/{(\%ile_{75}(X) - \%ile_{25}(X))}$, are \changed{preferred.}
				
				\paragraph{Frequency domain}
					A \ac{FFT} can be used to convert time domain data to the frequency domain, which can yield insights that are hidden in the time domain. However, it is not possible to track trends and patterns over time. Techniques such as cyclostationary analysis \cite{Randall2011} can help to analyze rotating machinery.
					
					The previously mentioned normalization techniques are applicable in this domain as well. To remove differences due to the unique machine characteristics, mechanical insights could be of use. For example, frequency spectra can be scaled by a specific frequency corresponding with a mechanical phenomenon or use signal processing techniques to obtain equal Parseval energy.
				
				\paragraph{Time/frequency domain}
					The time/frequency domain detects frequency trends and patterns over time and can be shown in a spectrogram. Different normalization procedures can be applied, depending on the application's need. One can normalize each of the obtained spectra using the techniques described in the frequency domain. Another option is to scale the complete spectrogram using the techniques from the time domain.
					
				\paragraph{Feature space}
					Various feature extraction methodologies, based on statistics, signal processing or physics, can extract indicators for fault diagnostics from any signal domain. Examples are amplitude in the time domain, order extraction or more complex signal processing techniques such as envelope analysis. The framework can use any of these indicators as a representation of machine behavior. In this case, normalization is required to equally weigh all features.
			
			\subsubsection{Dissimilarity measure}
				A measure\footnote{For the proposed framework implementation, a measure suffices. The triangle inequality, which must hold for a metric, is not required.} $s(X,Y)$ is needed to compare the distance or dissimilarity of a machine pair $(X, Y)$, represented by sequences $(x_1, x_2, \ldots, x_n)$ and $(y_1, y_2, \ldots, y_n)$ with length $n$. A data point $x_i$ or $y_i$ can be $1$-dimensional, such as for the time domain, or $m$-dimensional, for example when representing a spectrogram or a feature space with $m$ features.
				
				Various functions to measure dissimilarity between sequences can be considered. An existing one can be selected using engineering knowledge or one can be designed specifically for the problem at hand. Two popular measures are: 
			
				\paragraph{1. Euclidean distance}
					The Euclidean distance is the straight-line distance between two sequences of length $n$, given by:
					
					\begin{equation} \label{eq:block1_euclidean_distance}
					s(X, Y) = \sqrt{\sum_{i=1}^{n}{c(x_i, y_i)^2}}
					\end{equation}
					with $c(x_i, y_i)$ being the Euclidean norm between two $m$-dimensional data points: 
					\begin{equation} \label{eq:block1_euclidean_distanc_norm}
					c(x_i, y_i) = || x_i ||_2 - || y_i ||_2 = \sqrt{x_{i,1}^2 + \cdots + x_{i,m}^2} - \sqrt{y_{i,1}^2 + \cdots + y_{i,m}^2}
					\end{equation}
					A disadvantage of the Euclidean distance is its sensitivity to subtle shifts in the time series. This is solved by Dynamic Time Warping.
				
				\paragraph{2. Dynamic Time Warping}
					\ac{DTW} is a successful algorithm in time series analysis which aligns in a nonlinear fashion a sequence representing time series~\cite{Muller2007,Bemdt1994}. The goal is to find an alignment of sequences $X$ and $Y$ that has the minimal total distance. 
					The dissimilarity between $X$ and $Y$ is then expressed as:
					\begin{equation} \label{eq:block1_dtw}
					s(X, Y) = \min_{p}{\sqrt{\sum_{\theta=1}^{L} c\left(x_{p_{\theta,0}}, y_{p_{\theta,1}}\right)^2}}
					\end{equation}
					
					Where $p = (p_1, p_2, \ldots, p_L)$ expresses an optimal alignment with length $L$, also referred to as the warping path. Each $p_\theta$ represents a tuple where the first element refers to the element in the $x$ series on position $p_{\theta,0}$ and the second element refers to the element in the $y$ series on the $p_{\theta,1}$ position.
					It satisfies three conditions:
					\begin{enumerate}
						\item $p_1 = (1, 1)$ and $p_L = (n, n)$ (boundary condition)
						\item $p_{1,0} \leq p_{2,0} \leq \ldots \leq p_{L,0}$ and $p_{1,1} \leq p_{2,1} \leq \ldots \leq p_{L,1}$ (monotonicity condition)
						\item $p_{\theta+1} - p_\theta \in \{(1,0), (0,1), (1,1)\}$ for $1 \leq \theta \leq L-1$ (step size condition).
					\end{enumerate}
					
\begin{figure}
						\centering
						\begin{subfigure}[t]{.5\textwidth}
							\centering
							\includegraphics[width=\linewidth]{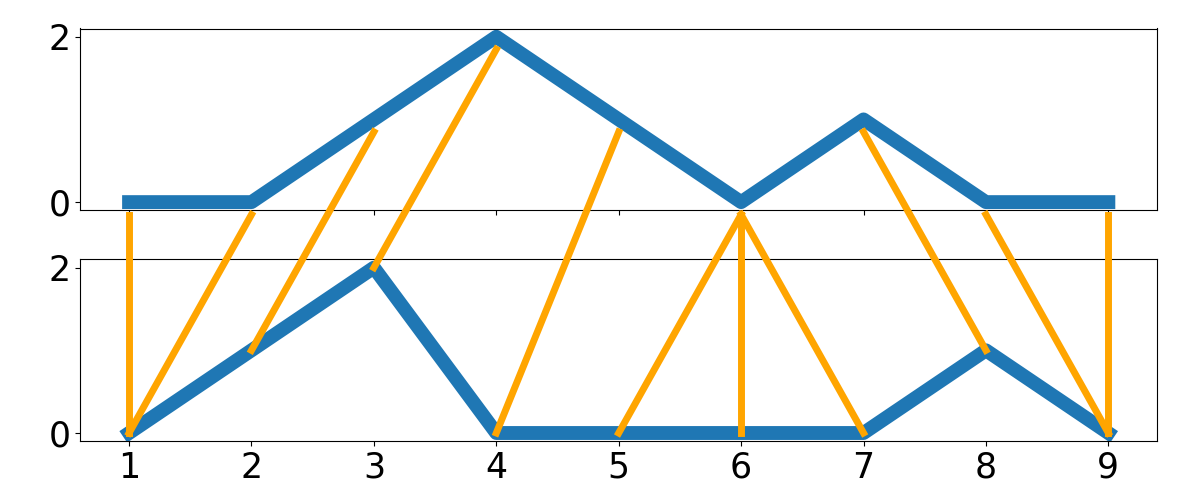}
							\caption{Original time series.}
							\label{fig:dtw_illustration_before}
						\end{subfigure}%
						\begin{subfigure}[t]{.5\textwidth}
							\centering
							\includegraphics[width=\linewidth]{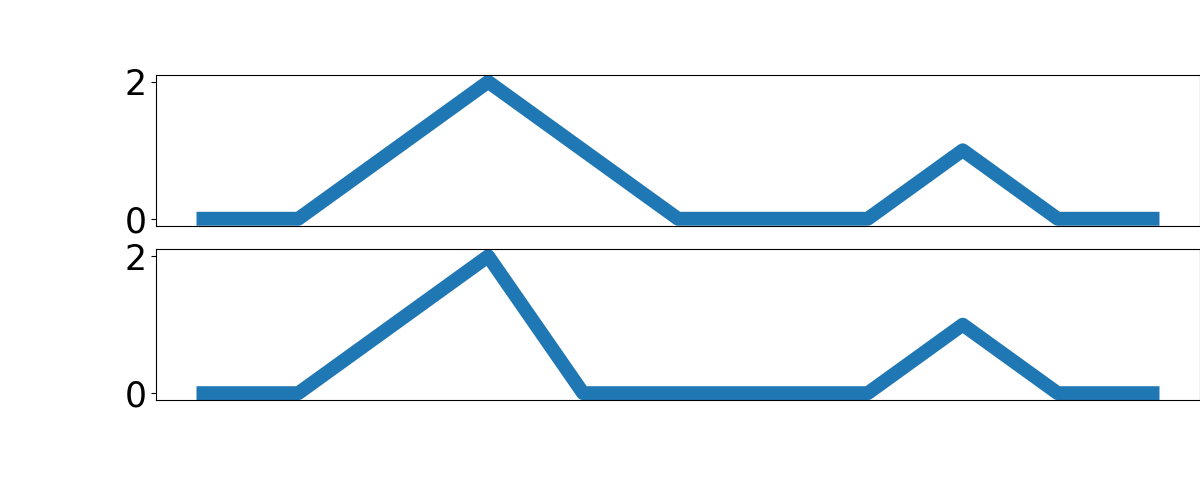}
							\caption{Optimal aligned time series. The X-axis is hidden, as the time domain is warped.}
							\label{fig:dtw_illustration_after}
						\end{subfigure}
						\caption{\ac{DTW} finds an optimal alignment between two time series (blue) by means of a warping path (orange). Euclidean distance in (\subref{fig:dtw_illustration_before}) is 7.0, the non-linear \ac{DTW} in (\subref{fig:dtw_illustration_after}) has a dissimilarity of 1.0.}
						\label{fig:dtw_illustration}
					\end{figure}

					The dissimilarity of individual data points $x_i$ and $y_j$ is defined by a cost function $c(x_i, y_j)$, the Euclidean norm (\cref{eq:block1_euclidean_distanc_norm}) by default. 
					The sequences  should be normalized before applying \ac{DTW}, to avoid suboptimal alignment due to differences in amplitude.
					\Cref{fig:dtw_illustration} shows an example in which two arbitrary time series are aligned. Note that the Euclidean distance is considerably larger compared to the \ac{DTW} dissimilarity even though the series are quite similar.
					
\begin{figure}
						\centering
						\begin{subfigure}[t]{.5\textwidth}
							\includegraphics[width=\linewidth]{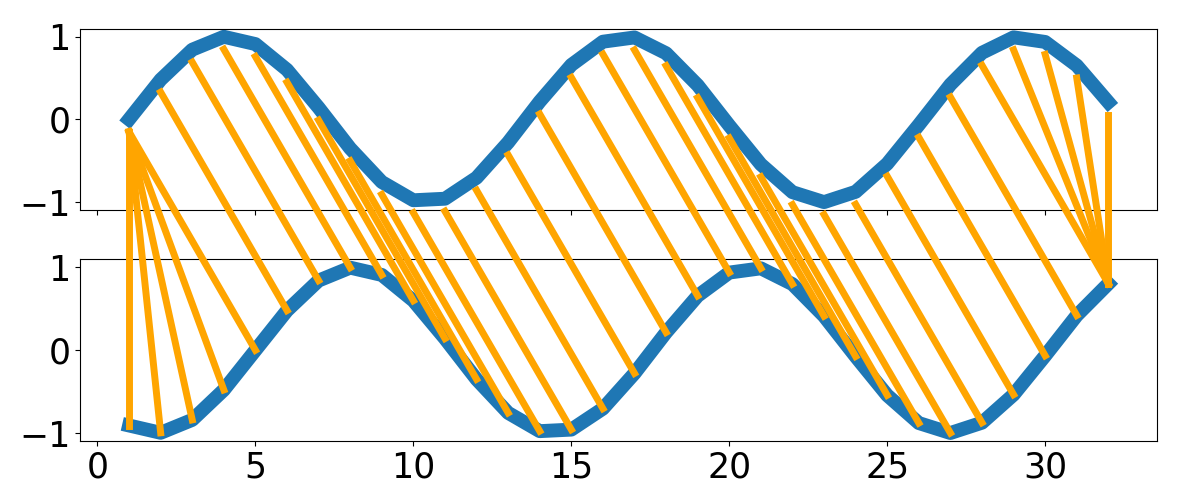}
							\caption{Classic \ac{DTW}. Dissimilarity: 1.8}
							\label{fig:dtw_relaxation_normal}
						\end{subfigure}%
						\begin{subfigure}[t]{.5\textwidth}
							\includegraphics[width=\linewidth]{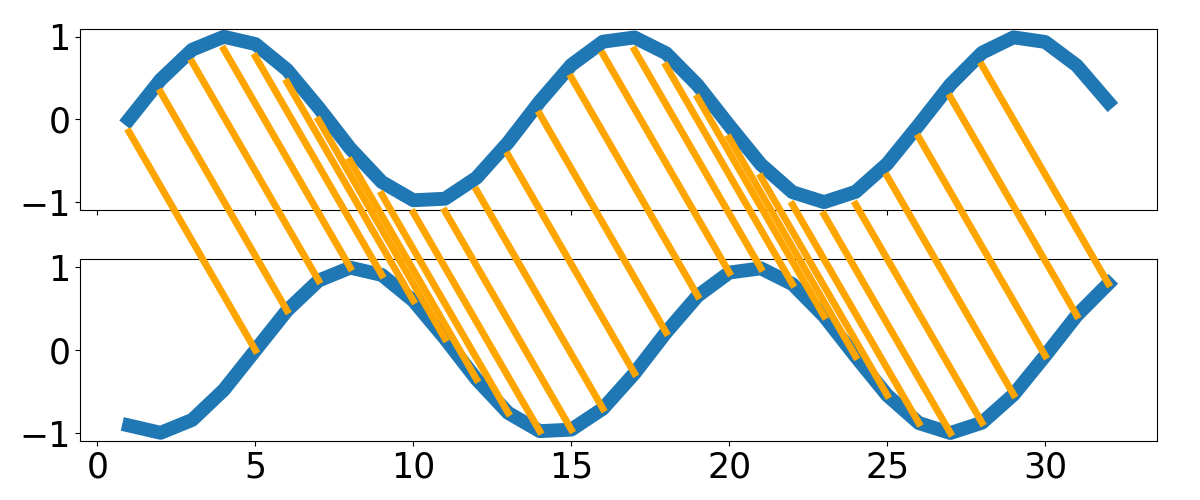}
							\caption{$\Psi$-\ac{DTW}. Dissimilarity: 0.0}
							\label{fig:dtw_relaxation_psi}
						\end{subfigure}
						\caption{\ac{DTW} on sine waves with a phase shift. Warping paths are shown in orange. While identical, the sine waves are considered dissimilar by classic \ac{DTW} due to additional warping in the beginning and end. $\Psi$-\ac{DTW} removes this effect and correctly identifies the sine waves as identical. }
						\label{fig:dtw_relaxation}
					\end{figure}

				A variant to DTW used in this work is $\Psi$-\ac{DTW}.
					Periodic signals can be misaligned due to a phase offset, which can result in \ac{DTW} introducing undesired warping at the beginning and end (\cref{fig:dtw_relaxation_normal}). This additional warping can lead to an unwanted increase in the \ac{DTW} measure.
					$\Psi$-\ac{DTW}~\cite{Silva2016} addresses this issue and avoids the undesired warping (\cref{fig:dtw_relaxation_psi}). It does this by relaxing the \ac{DTW} boundary condition by $p_1 = ([1,\Psi], 1)$ or $p_1 = (1, [1,\Psi])$, and $p_L = ([n-\Psi,n], n)$ or $ p_L = (n, [n-\Psi,n])$, where $\Psi \in \mathbb{N}$ is a user-defined parameter.
					Apart from this, it operates in a similar way as the classic \ac{DTW}. The example in \cref{fig:dtw_relaxation} shows that for periodic signals, $\Psi$-\ac{DTW} leads to a better match.
					
				\subparagraph{\ac{DTW} warping amount}
					$(\Psi$-)\ac{DTW} is known to work well when it is important to compare time series based on their shape. However, in some cases, \ac{DTW} removes the difference between healthy and faulty signals. One example is in electrical current waveforms which are considered in one of the use-cases.
					To overcome this challenge, a measure that considers the amount of warping that occurs in $\Psi$-\ac{DTW} is used \cite{Hendrickx2019}:
					
					\begin{equation} \label{eq:block1_dtwwarping}
					s(X, Y) = \sum_{\theta=1}^{L-1} [p_{\theta+1} - p_{\theta} \neq (1,1)].
					\end{equation}
					It is advised to use a normalized form, $s(X,Y)/L$, when using $\Psi$-\ac{DTW}, as the length of the paths can vary.
			
		\subsection{Building block 2: Clustering the machines in the fleet}
			\label{sec:fleet_clustering}
			This block identifies similarly behaving machines by employing a clustering algorithm that uses the selected measure $s(X,Y)$ to compare pairs of machines. There are two important points to consider when selecting a clustering algorithm. First, the selected clustering algorithm should work for an arbitrary measure $s(X, Y)$. Some algorithms only work with certain measures (e.g., \textit{K-means} requires a space where the mean is defined). Therefore, it is preferable to use an algorithm that represents clusters by actual data points and is hence more generic, such as K-medoids \cite{Kaufman1990,Park2009}, spectral clustering \cite{VonLuxburg2007}, hierarchical clustering \cite{Murtagh1983} and DB-scan \cite{Ester1996}.
			Second, the number of clusters is unknown in anomaly detection. If all the machines are healthy, a single cluster is expected whereas multiple clusters are needed if there are faulty machines. Some methods, like K-medoids, require the user to specify the number of clusters $k$ in advance. In contrast, approaches like hierarchical clustering and DB-scan derive the number of clusters from the structure of the data, often by using a parameter that describes how tight a cluster should be. 
			
			Hierarchical clustering is proposed as the default implementation for this block. It offers the benefits of being relatively simple, producing results that can be easily visualized, and not having to set $k$ upfront.
			
			\paragraph{Hierarchical clustering}
				Hierarchical clustering builds a hierarchy of clusters. This can be done in either an agglomerative (bottom-up) or divisive (top-down) manner. However, agglomerative clustering is more common. This approach begins by assigning each item to its own subcluster. Then, the algorithm iteratively merges the two nearest (or most similar) subclusters until there is only one cluster left.  The function to measure the distance between two subclusters is a parameter of hierarchical clustering. Two common choices are: \textit{single linkage}, which considers the minimum distance between the elements of two subclusters and \textit{complete linkage}, which considers the maximum.  Other alternatives are \textit{average linkage} and Ward's method \cite{Ward1963}.
				
\begin{figure}
					\centering
					\includegraphics[width=0.4\textwidth]{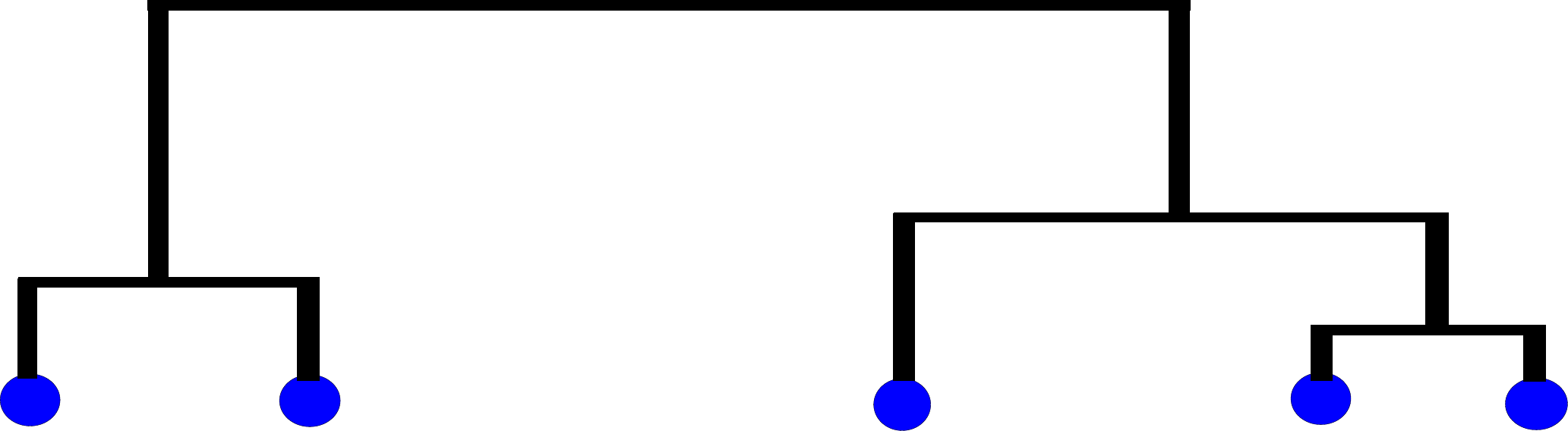}
					\caption{A dendrogram representing a cluster hierarchy. Agglomerative (bottom-up) clustering begins by assigning each item (blue dot) to its own cluster and iteratively merges the two most similar clusters until only one cluster is left. In this example, the horizontal distance between the items is considered being their distance.}
					\label{fig:dendrogram}
				\end{figure}

				The result of the clustering can be represented in a tree-like structure called a dendrogram (\cref{fig:dendrogram}). It is relatively easy to interpret this structure, which is an additional benefit over other clustering techniques. At the lowest level, each cluster contains a single item. Going from the bottom-up, each subcluster is merged with its most similar subcluster until a single cluster is obtained. The vertical axis shows the distance between each of these subclusters. The higher this merge occurs, the less similar the subclusters are.
				
				The proposed anomaly detection block requires that each machine is assigned to only one cluster. The hierarchical structure of the clustering needs to be partitioned to satisfy this constraint. A naive approach is to cut the hierarchy by using a threshold on the distance between subclusters. Selecting an appropriate threshold depends on $s(X, Y)$ (e.g., its range)  and does not take the underlying cluster structure into account. To overcome this challenge, a threshold on the cophenetic correlation $thr_{cc}$ is proposed. This offers the advantages of being independent of the range of $s(X, Y)$ and considering the structure of the data.
				
				The cophenetic correlation measures how well a clustering preserves the original pairwise distances \cite{Lessig1972}, that is, how well the clustering reflects the original structure of the data. A high cophenetic correlation indicates that two or more cluster partitions are present in the data, while a low cophenetic correlation suggests that only a single cluster partition is present in the data. Formally, the cophenetic correlation is defined as \cref{eq:block2_cophenetic_correlation}:
				\begin{equation} \label{eq:block2_cophenetic_correlation}
				cophenetic\_correlation = \frac {\sum_{(X,Y) \in P} (s(X,Y) - \bar{s})(t(X,Y) - \bar{t})}{\sqrt{[\sum_{(X,Y) \in P}(s(X,Y)-\bar{s})^2] [\sum_{(X,Y) \in P}(t(X,Y)-\bar{t})^2]}}.
				\end{equation}
				where $P$ is the set of all machine pairs $(X, Y)$, $s(X, Y)$ is the measure used in the clustering, and $t(X, Y)$ is the dendrogrammic distance, which is the height at which machines $X$ and $Y$ are first joined in the dendrogram.\footnote{Note that the dendrogrammic distance depends on the selected linkage function.} Finally, $\bar{s}$ and $\bar{t}$ are the average pairwise dissimilarities and dendogrammic distances. A pseudocode for splitting a dendrogram is given in  \cref{algo:partition_dendrogram}.
				
\begin{figure}
					\centering
					\includegraphics[width=\textwidth]{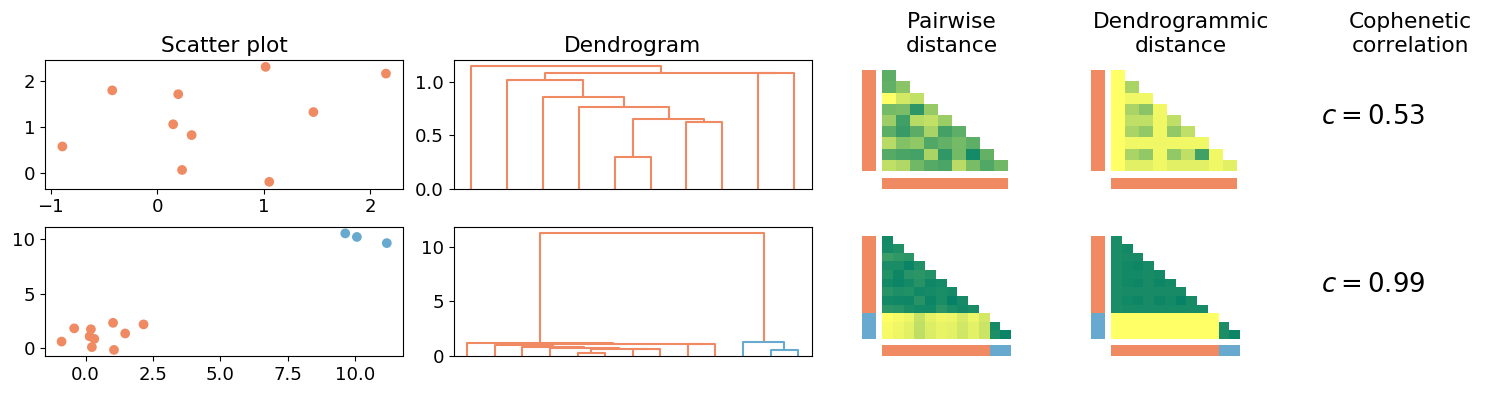}
					\caption{Upper: 10 data points form one cluster. Single linkage hierarchical cluster generates the corresponding dendrogram. There is a clear difference between pairwise and dendrogrammic distances, shown in triangular matrices with a green (low) to yellow (high) color scale. This difference leads to a low correlation.\\
						Lower: a second cluster (blue) of 3 additional data points. In this situation, the difference in pairwise and dendrogrammic distances is relatively small. This leads to a higher correlation, indicating the presence of multiple clusters.}
					\label{fig:copehentic_correlation}
				\end{figure}

				The intuition behind the cophenetic correlation procedure is illustrated with a simple example (\cref{fig:copehentic_correlation}). First, the upper part considers a single cluster (shown in red). There is a difference between the pairwise and dendrogrammic distances. The correlation between them is rather low. Second, the lower part of the figure shows a situation where a second cluster is present (shown in blue). Now, the distances within a cluster are relatively small. In this situation, the larger distances between clusters dominate the correlation, which is now close to 1.
				
				\begin{algorithm}
					\caption{Hierarchical cluster partitioning}\label{algo:partition_dendrogram}
					\begin{algorithmic}[1]
						\Procedure{Partition($dendrogram$, $thr_{cc}$)}{}
						\State $cc = cophenetic\_correlation(dendrogram)$ (\cref{eq:block2_cophenetic_correlation})
						\If {$cc > thr_{cc} $}
						\State $d_1$, $d_2$ = cut\_at\_highest\_level(dendrogram)
						\State \Return Partition($d_1$, $thr_{cc}$), Partition($d_2$, $thr_{cc}$)
						\Else {
							\State \Return d
						}
						\EndIf
						\EndProcedure
					\end{algorithmic}
				\end{algorithm}
				
				The cophenetic correlation threshold $thr_{cc}$ is the most important parameter in the default implementation, as it controls the cluster partitioning. It is set depending on the cost of not identifying a faulty machine (i.e., a false negative, missed detection) or incorrectly labeling a healthy machine as faulty (i.e., a false positive, false alarm). \changed{Moreover, it determines when an alarm occurs when monitoring slowly degrading machines.} If $thr_{cc}$ is set too high, a faulty machine might be partitioned together with \changed{machines considered to be healthy}. The anomaly detection procedure would then report faulty machines as being healthy. Multiple small partitions containing \changed{machines considered to be healthy} would be formed if $thr_{cc}$ is set too low. Anomaly detection will then report \changed{these} machines as being faulty. The value of this threshold is set depending on the application.
		
		\subsection{Building block 3: Anomaly detection}
			\label{sec:anomaly_score}
			Block 3 uses the discovered clustering to assign an anomaly score to each machine. Machines with similar behavior are expected to be in the same cluster. The larger a cluster, the more likely it is to indicate healthy behavior. Each machine's anomaly score is simply the fraction of total machines that are \changed{not} assigned to its cluster\changed{, hence an anomaly score close to one indicates a machine that is likely to be faulty.} Thresholding this score to classify each machine as either healthy or faulty requires reasoning about the size of the fleet and the expected proportion of faulty machines.
			
			More sophisticated unsupervised anomaly and outlier detection techniques can be used. These can be based on clustering, neighborhood, or statistics \cite{Pimentel2014,Chandola2009,Agrawal2015}. Cluster-based techniques are preferred for fleet-condition monitoring applications. This class of techniques identifies anomalies as small clusters or unclustered data points. Neighborhood-based techniques like \ac{kNN} or \ac{LOF} detect anomalies based on respectively the distance to the K-th nearest neighbor or the density of its neighborhood. However, the accuracy of these methods depends on the number of analyzed machines, which can be limited. Statistical-based techniques assume that anomalies arise in low-probability regions. However, selecting an appropriate hypothesis test for high-dimensional data is non-trivial. Moreover, these tests require making assumptions about the underlying data distribution, which might not hold in practice.
		
		\subsection{Building block 4: Visualization}
			\label{sec:visualization}
\begin{figure}
				\centering
				\includegraphics[width=0.8\textwidth]{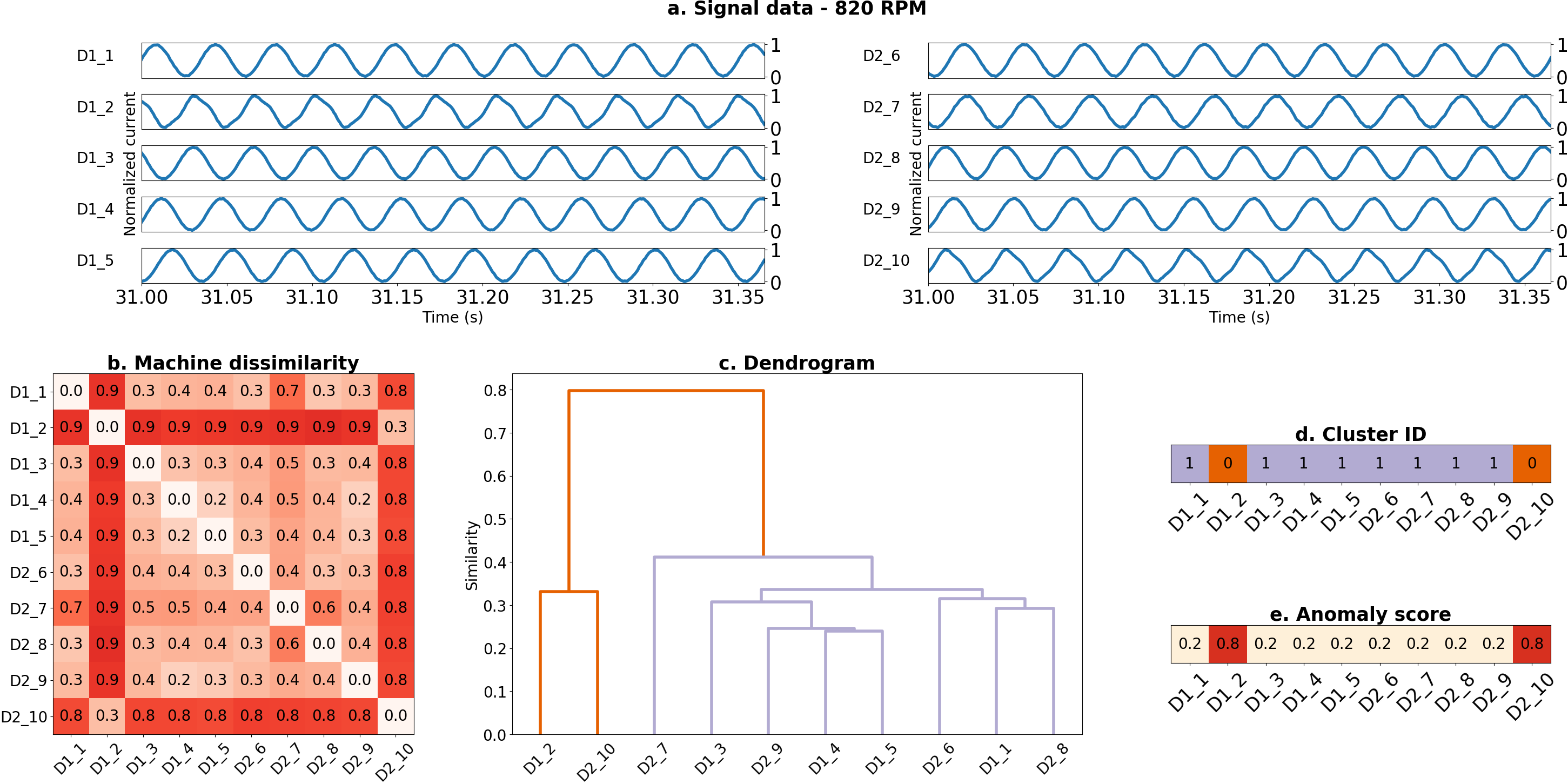}
				\caption{An example of the \changed{visualizations provided by the} framework, which are analyzed in detail in \cref{sec:uc1_var1}\removed{ (fig. 12)}. In this example, drivetrains D1\_2 and D2\_10 are faulty.\removed{ For convenience, this image is shown twice.} Signal data (a.) gives the user a detailed view. \changed{In this example, min-max normalization has been applied to the waveforms.} A global overview of the fleet is given by pairwise comparisons (b.) and a dendrogram (c.). Finally, the cluster partitions (d.), \changed{obtained with $thr_{cc} = 0.9$, } and their corresponding anomaly scores (e.) indicate \changed{the prediction for a} machine's health condition\removed{ predictions}. \changed{In this example, the framework correctly identifies D1\_2 and D2\_10 as being anomalous.}
				}
				\label{fig:results_template}
			\end{figure}

\begin{figure}
				\centering
				\includegraphics[width=\textwidth]{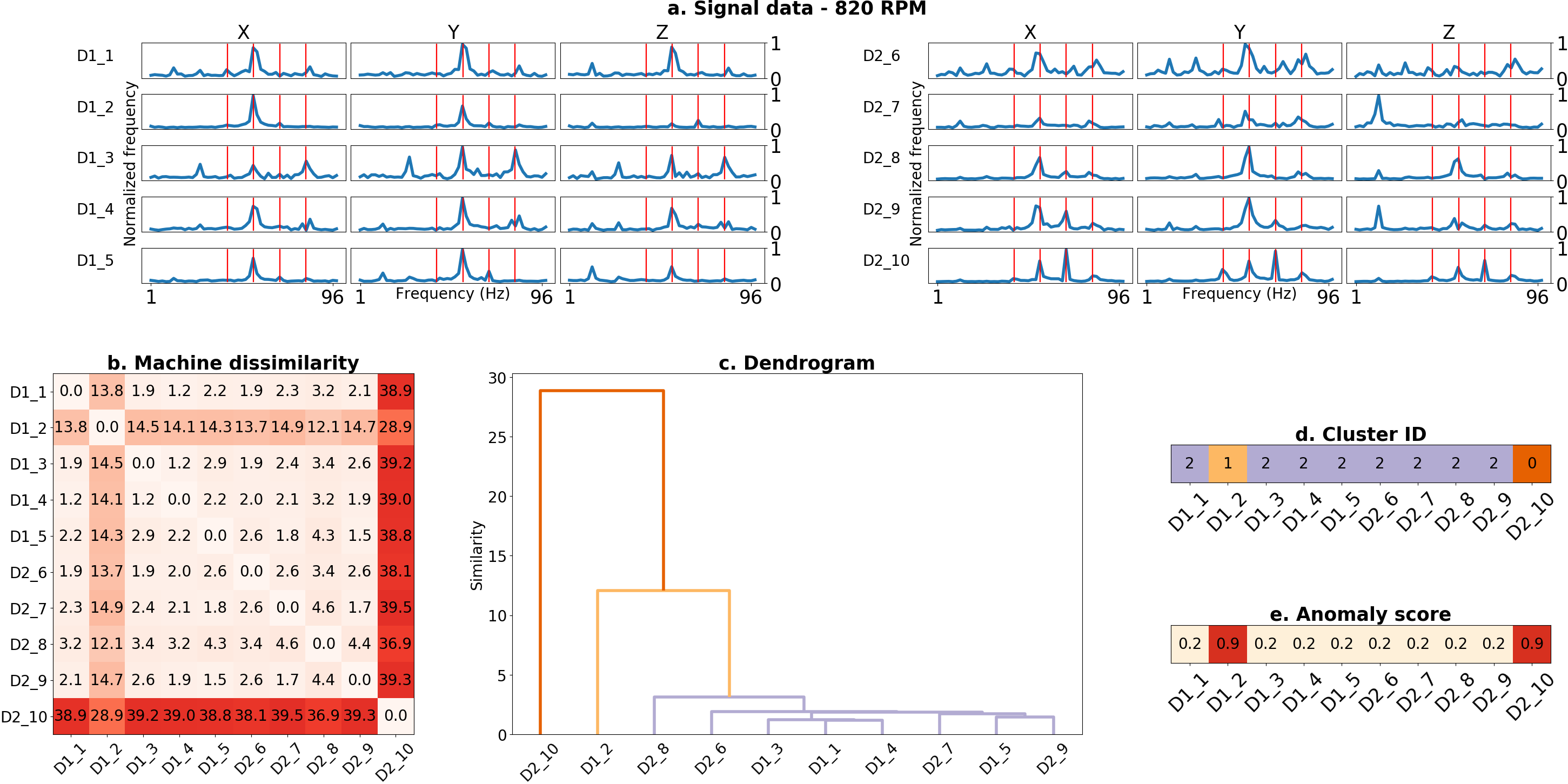}
				\caption{Visualization showing the three accelerometer frequency domain signals (a.), machine similarities (b.), clustering (c. \& d.) and anomaly scores (e.). The red vertical lines in (a.) indicate the used harmonics 3 -- 6. Clusters are partitioned with $thr_{cc} = 0.9$. Each machine is running stationary at 820 RPM, with D1\_2 and D2\_10 having a voltage unbalance. D1\_2 is less affected by the voltage unbalance compared to D2\_10, which is visible in the dendrogram.}
				\label{fig:result_pu_acc_freq_35}
			\end{figure}

			Visualizing the results of each block can yield insights into the machine health conditions and the framework's predictions. A simple visualization  shows results of each block (e.g. \cref{fig:results_template}). The use-cases show more examples in \cref{fig:result_pu_curr_freq_35,fig:result_pu_curr_timefreq_35,fig:result_pu_acc_freq_35}.
			
			First, both parts of the machine comparison block can give detailed insights. The template has space for preprocessed machine data, to give the user a more detailed view (\cref{fig:results_template}a.). A dissimilarity matrix (\cref{fig:results_template}b.) shows the result of pairwise machine comparisons. A color map helps to see relative differences, as absolute values depend on the selected comparison method.
			Second, clustering gives a more global overview of the machine fleet. A dendrogram can, for example, show a cluster hierarchy, to detect divergent machines (\cref{fig:results_template}c.). Moreover, colors can indicate cluster partitions (\cref{fig:results_template}d.). This helps to tune the threshold $thr_{cc}$. Finally, the anomaly scores correspond to the predicted health conditions (\cref{fig:results_template}e.).
			
	\section{Application and evaluation of the proposed framework on experimental use-cases}
		\label{sec:use_cases_general}
		The proposed framework is validated in two different use-cases. Both involve the detection of a voltage unbalance in a fleet of electrical drivetrains. This results in early machine failure and is often caused by a grid fault or a bad wire connection. Early detection will extend the lifetime of a machine and will minimize wear due to increased temperature and additional vibrations. 
		
		Each use-case includes signal processing techniques for condition monitoring. These are used as a benchmark for the framework. Its broad applicability is shown by different variants, using different degrees of domain knowledge. Each variant makes use of different implementations for machine comparison but uses the default fleet clustering, anomaly detection, and visualization blocks. It analyzes and predicts machine status for 0.5s non-overlapping analysis windows. A machine is considered to be faulty after being identified as anomalous for five consecutive windows. This reduces misclassifications due to signal noise. 
		
		Three performance metrics consider the two types of misclassifications: false negatives (fn) and false positive (fp) (\cref{tab:prediction_mistakes}). Precision (\changed{\cref{eq:precision}}) and Recall (\changed{\cref{eq:recall}}) help the user to take the cost into account of labeling a healthy machine as being faulty (precision, false alarm) versus missing a faulty machine (recall, missed detection). The F1-score (\changed{\cref{eq:f1}}) considers both faults equally important and is used as a single performance measure. 
		
		The evaluation procedure considers both stationary as well as dynamic operational conditions. The former allows analyzing the consistency of the framework, the latter its generality. Within these operational conditions, the health status of machines D1\_2 and D2\_10 is varied. 
		
		\begin{table}[h!]
			\caption{Possible prediction mistakes}
			\label{tab:prediction_mistakes}
			\begin{tabular}{lllll}
				&         & \multicolumn{2}{c}{\textbf{True condition}}  &  \\
				&         & \textbf{Faulty}          & \textbf{Healthy}         &  \\
				\multirow{2}{*}{\textbf{Predicted condition}} & \textbf{Faulty} & True positive (tp)  & False positive (fp) & \changed{\textbf{Precision} (\cref{eq:precision})} \\
				& \textbf{Healthy}  & False negative (fn) & True negative (tn) & \\
				& & \changed{\textbf{Recall} (\cref{eq:recall})} & & \changed{\textbf{F1-score} (\cref{eq:f1})}
			\end{tabular}
		\end{table}
	
	\changed{
		\begin{subequations}
			\begin{equation} \label{eq:precision}
				Precision = \frac{tp}{tp+fp}
			\end{equation}
			
			\begin{equation} \label{eq:recall}
				Recall = \frac{tp}{tp+fn}
			\end{equation}
			
			\begin{equation}
				\label{eq:f1}
				F1-score = 2 \frac{Precision . Recall}{Precision + Recall}
			\end{equation}
		\end{subequations}
	}
	
	\section{Use-case 1: Voltage unbalance -- Electrical signatures}
		\label{sec:use_case_vu_curr}
		This use-case involves electrical measurements of an electrical machine fleet. The task is to detect a voltage unbalance fault based on current signals. Three variants show the general applicability of the framework in different signal domains. The first variant uses time domain data and demonstrates a generic approach for waveform analysis. The second variant is an example on how to include domain knowledge. Finally, the third variant is a hybrid method and analyzes time/frequency data.
		
		\subsection{Experimental fleet test rig and data set}
			\label{sec:uc1_data}
\begin{figure}[h]
				\centering
				\begin{subfigure}[t]{.5\textwidth}
					\centering
					\includegraphics[width=0.9\linewidth]{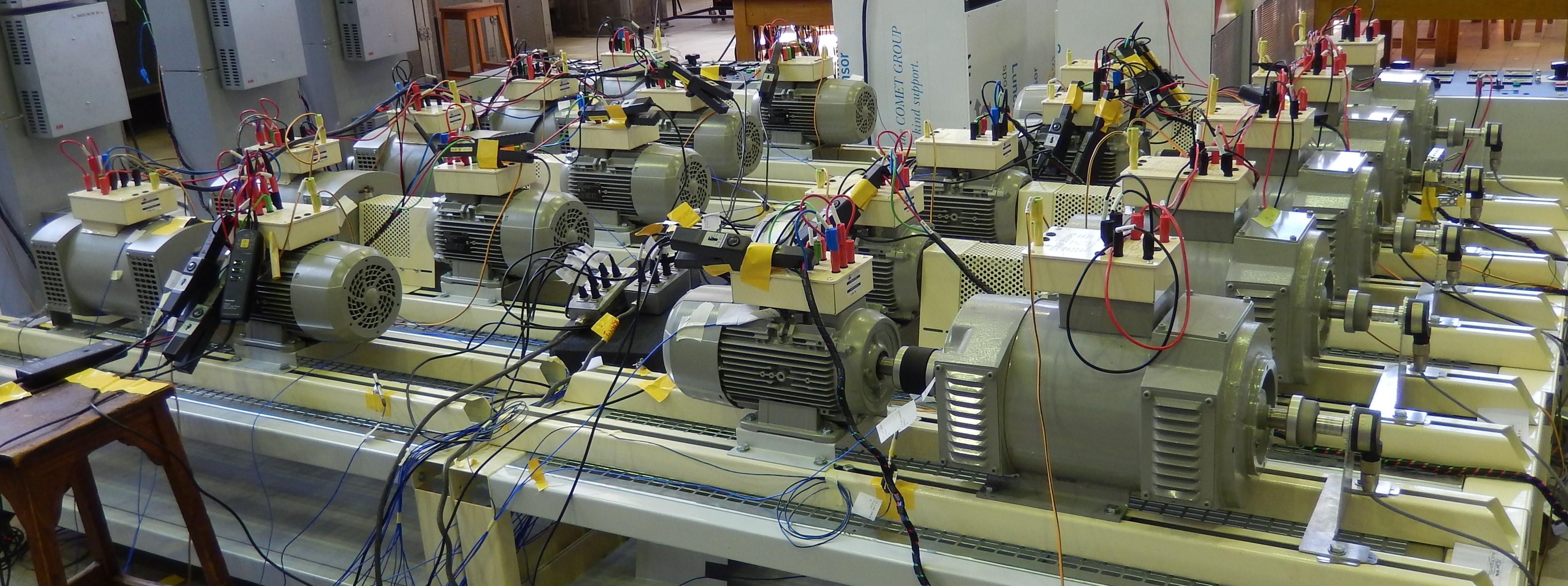}
					\caption{Motor fleet.}
					\label{fig:data_setup}
				\end{subfigure}%
				\begin{subfigure}[t]{.5\textwidth}
					\centering
					\includegraphics[width=\linewidth]{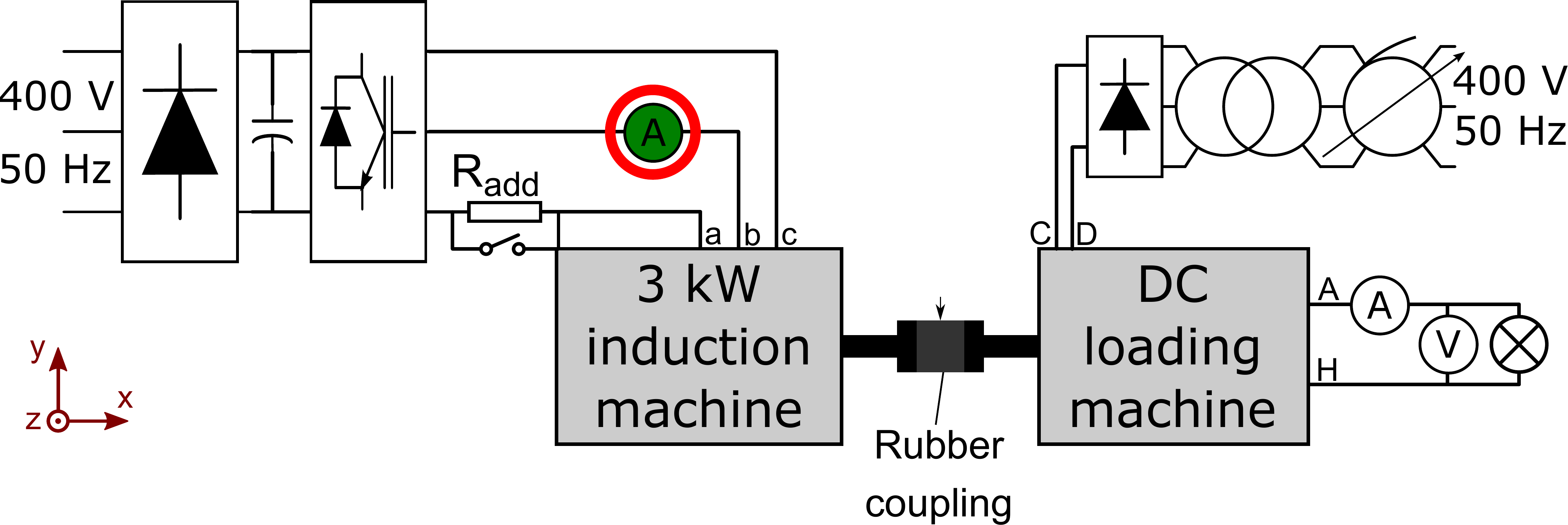}
					\caption{Schematic of drivetrain D1\_2. The analyzed current sensor (green) measures phase B.}
					\label{fig:ULB_bench_curr}
				\end{subfigure}%
				\caption{The measurement setup consists of ten drivetrains, having a 3kW \ac{SCIM} (drive, inner part of the fleet) coupled to a 3kW \ac{WRSM} (left) or \ac{DC} (right) motor (load). Drive and load side motors are connected with a rubber coupling.}
				\label{fig:ULB_bench_curr_setup}
			\end{figure}

			A single phase current of ten electrical drivetrains (\cref{fig:ULB_bench_curr_setup}) is measured at 25600 Hz. Each drivetrain consists of an electrical motor pair, connected by a flexible jaw coupling. A 3-phase \ac{SCIM}\footnote{Siemens 1AV3105B 1LE10031AB522AA4} drives the shaft. Ten ABB\footnote{ABB ACS880-01-09A4-3} drive controllers manage the drive sides with internal closed-loop direct-torque control. All drivetrains are set up to always have similar speeds. The other motor acts as load and is either a 3 kW \ac{DC} motor (drivetrains D1\_1 -- D1\_5) or a 3 kW \ac{WRSM} (drivetrains D2\_6 -- D2\_10). The specifications of each machine are shown in \cref{tab:drive_side_config}. Resistors generate load while keeping the current excitation constant. The resulting load torque is proportional to speed and corresponds to the rated load at rated speed.
			
			A \SI{3}{\ohm} external resistor $R_{add}$ emulates a voltage unbalance in drivetrains D1\_2 and D2\_10. This resistor is inserted between the drive controller and the motor (\cref{fig:ULB_bench_curr}). A higher resistance would trigger the internal safeties of the ABB drive controller.
			
			The data set includes both stationary and dynamic system behavior under loaded and unloaded operating conditions. For the stationary case, each drivetrain is running at a fixed speed (820 RPM or 1500 RPM). On the other hand, during the dynamic operational condition, all machines perform a run-up from 0 RPM to 1200 RPM.
			
			\begin{table}[h!]
				\centering
				\caption {Rated parameters of motors}
				\label{tab:drive_side_config}
				\begin{tabular}{| l | l | l | l | l | l | l |}
					\hline
					& Power (kW) & Phase-to-phase voltage (V) & Current (A) & Frequency (Hz) & Pole pairs & Speed \\ \hline
					\ac{SCIM} & 3 & 400 & 6.2 & 50 & 2 & 1385 \\ \hline
					\ac{WRSM} & 3 & 400 & 4.3 & 50 & 2 & 1500 \\ \hline
					\ac{DC} & 3 & 270 & 11.2 & DC & 1 & 1500 \\ \hline
				\end{tabular}
			\end{table}

		\subsection{Classic voltage unbalance diagnosis techniques}
			\label{sec:pu_curr_sota}
			\ac{MCSA} is a popular class of techniques for detecting motor faults that apply signal processing methods to current signals. It can be used to detect a voltage unbalance~\cite{Sharifi2011,Mollet2018}. Some of these methods require measuring multiple motor current signals. For example, the phase shift between the three current signals of a healthy system is 120 degrees. A voltage unbalance results in changes to this angle \cite{Sang-JoonLee}. Another example measures an estimator for $I_{RMS}$ based on two\footnote{The third current signal can be inferred from the measured signals.} current signals. A fluctuating component at twice the fundamental frequency is observed due to the presence of inverse sequence current in the stator \cite{Mollet2018}.
			
\begin{figure}
				\centering
				\begin{subfigure}[t]{.5\textwidth}
					\centering
					\includegraphics[width=\linewidth]{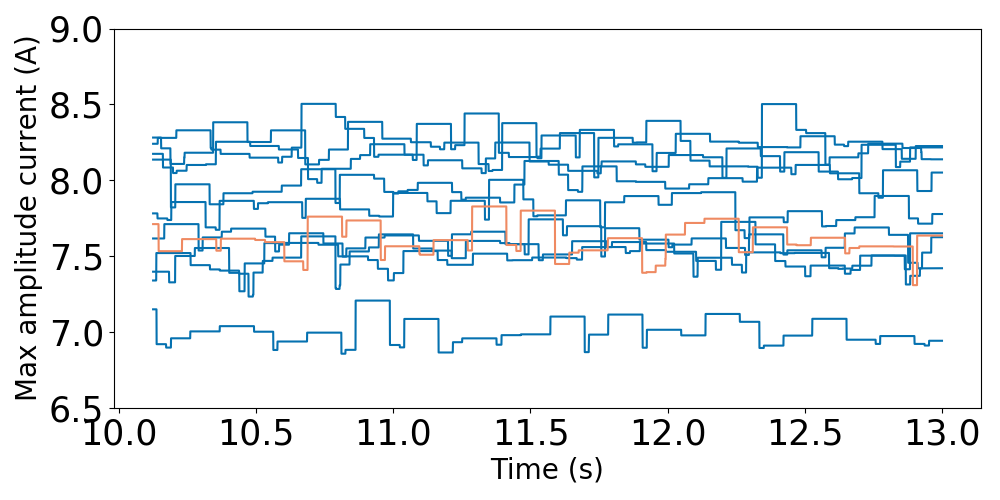}
					\caption{All machines healthy}
					\label{fig:intermachine_difference_0}
				\end{subfigure}%
				\begin{subfigure}[t]{.5\textwidth}
					\centering
					\includegraphics[width=\linewidth]{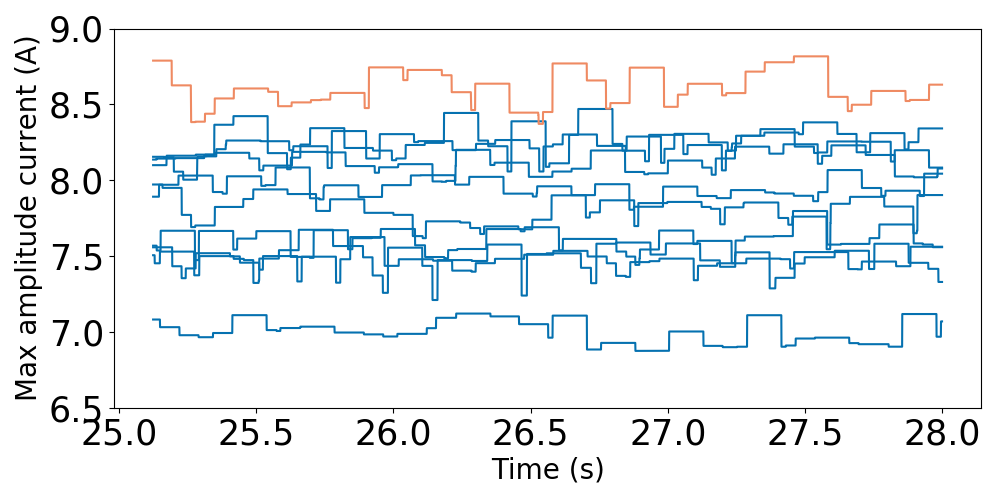}
					\caption{D2\_10 (orange) faulty}
					\label{fig:intermachine_difference_1}
				\end{subfigure}
				\caption{Voltage unbalance detection based on amplitude is not possible in a fleet context. Machine D2\_10 is indicated in orange. The variance within the healthy fleet is larger than its amplitude increase due to a voltage unbalance. All machines are running stationary at 820 RPM (loaded).}
				\label{fig:intermachine_difference}
			\end{figure}

\begin{figure}
				\centering
				\begin{subfigure}[t]{.33\textwidth}
					\centering
					\includegraphics[width=\linewidth]{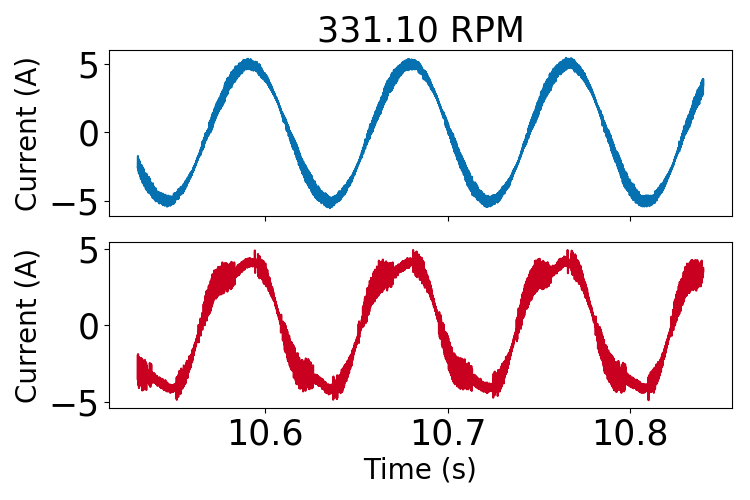}
				\end{subfigure}%
				\begin{subfigure}[t]{.33\textwidth}
					\centering
					\includegraphics[width=\linewidth]{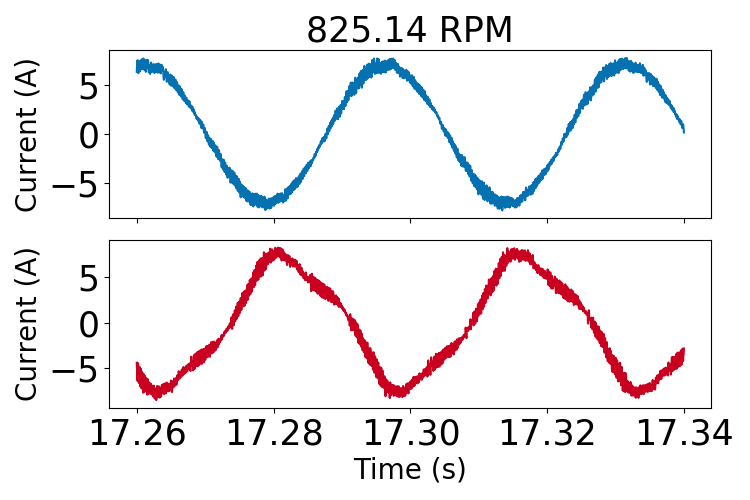}
				\end{subfigure}%
				\begin{subfigure}[t]{.33\textwidth}
					\centering
					\includegraphics[width=\linewidth]{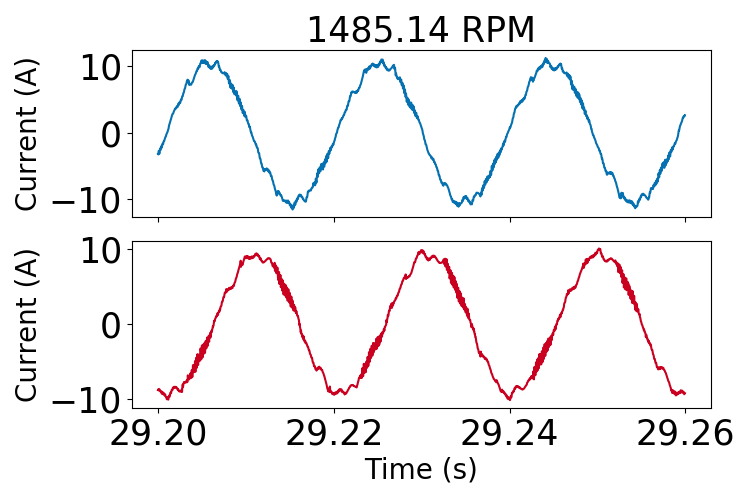}
				\end{subfigure}
				\caption{Current waveform analysis can detect a voltage unbalance. The fault affects the current waveform of a faulty drivetrain (red). The exact shape is however \changed{dependent} on the operational speed. Above 1385 RPM, the fault effect disappears due to an increase of the flux. A current signal of a healthy drivetrain (blue) is shown for reference.}
				\label{fig:ds1_time_effect}
			\end{figure}

\begin{figure}
				\centering
				\begin{subfigure}[t]{.33\textwidth}
					\centering
					\includegraphics[width=\linewidth]{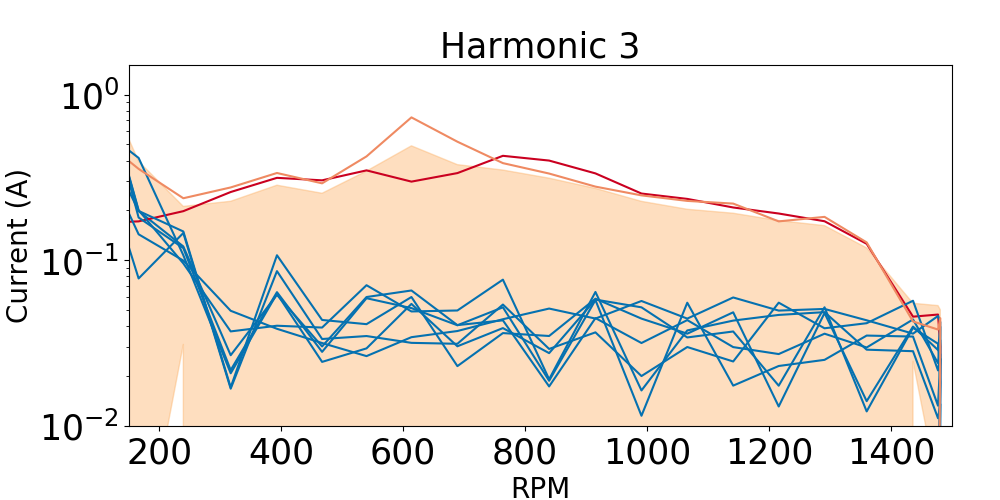}
					\caption{Load, D1\_2 \& D2\_10 faulty.}
					\label{fig:ds1_order_runup_load}
				\end{subfigure}%
				\begin{subfigure}[t]{.33\textwidth}
					\centering
					\includegraphics[width=\linewidth]{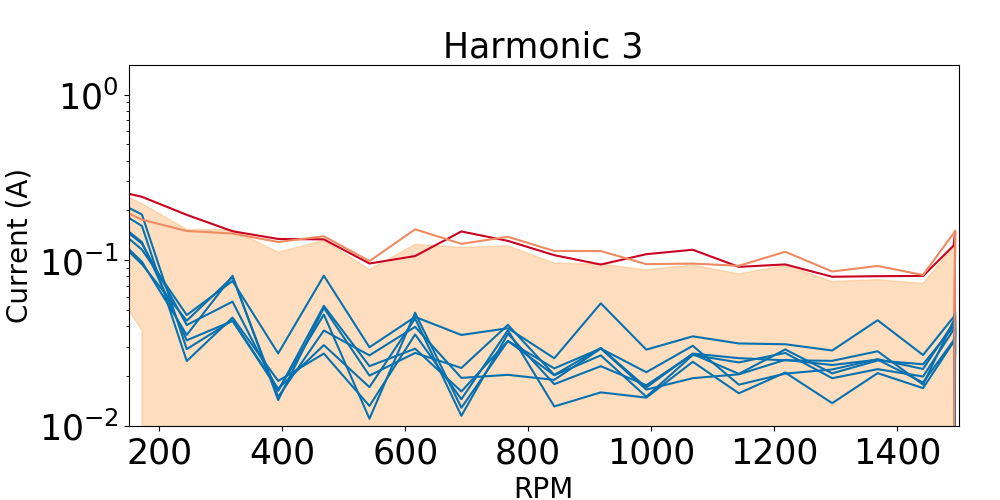}
					\caption{No load, D1\_2 \& D2\_10 faulty.}
					\label{fig:ds1_order_runup_noload}
				\end{subfigure}%
				\begin{subfigure}[t]{.33\textwidth}
					\centering
					\includegraphics[width=\linewidth]{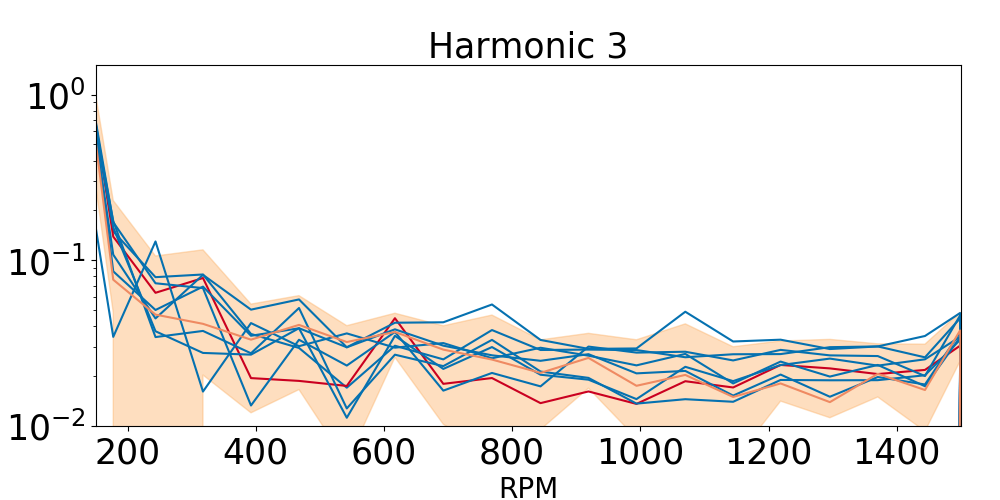}
					\caption{No load, all machines healthy.}
					\label{fig:ds1_order_runup_noload_healhty}
				\end{subfigure}
				\caption{The third harmonic can be used to detect a voltage unbalance. Faulty machines D1\_2 (red) and D2\_10 (orange) show a clear increase compared to the healthy (blue) machines between 300 and 1420 RPM. The orange band indicates the range in which a machine is considered healthy, being 2 standard deviations around the mean. This correctly detects the machine health condition in most of the cases (\cref{tab:pu_curr_harmonics}).}
				\label{fig:ds1_order_runup}
			\end{figure}

			A single current sensor is sensitive to a secondary effect of the voltage unbalance \cite{Mollet2018}. The oscillations of the shaft lead to an increase in the third harmonic. This work compares the framework with three approaches that make use of this signal.
			First, careful time domain analysis reveals an increase in the current amplitude. However, a fleet-based approach cannot use this current amplitude as the difference due to a fault is lower than the variability within the fleet (\cref{fig:intermachine_difference}). Moreover, the current amplitude highly depends on the operational conditions (i.e. rotational speed and load).
			Second, the fault affects the current waveform. However, the shape of the waveform depends on the operational speed (\cref{fig:ds1_time_effect}). Any type of deviation should thus trigger a machine fault prediction.
			Finally, the frequency domain shows an increase of the third harmonic amplitude (\cref{fig:ds1_order_runup}). The harmonic amplitude is not sensitive to changes in speed but does depend on the load. Different load conditions could thus require different thresholds. 
			
			In each of these approaches, the machine is affected up to a certain speed ($\sim 1400 $RPM). This is demonstrated in \cref{fig:ds1_order_runup}, where the amplitude of the third harmonic decreases at that point. The hypothesis is that at higher speeds, a flux weakening region is entered. For example at the rated load above 1385 RPM, the stator frequency exceeds the rated 50 Hz. Therefore, the flux is decreased in the machine to keep the stator voltage under its limit. As a consequence, the magnetic saturation level decreases and the third harmonic in the current becomes less visible.
			
			The remainder of this use-case utilizes a simple condition monitoring method as a benchmark to validate the proposed framework. This method considers a machine as healthy when its third harmonic amplitude is within $\sigma$ standard deviations of the fleet's mean. Note that this is a simplified fleet approach, which avoids setting thresholds manually. \Cref{tab:pu_curr_harmonics} shows the fault detection performance for this methodology with different values of $\sigma$, \Cref{fig:ds1_order_runup} illustrates the specific case for $\sigma = 2$. 
			
			\begin{table}
				\caption{Performance of classic voltage unbalance detection using the third current harmonic. Scores corresponding to a scenario's highest F1-score are indicated in bold.}
				\label{tab:pu_curr_harmonics}
				\begin{tabular}{ l  l r r r r r r r r  }
    & & \multicolumn{ 8 }{ c }{ $\sigma$ } \\
    & Metric & 0.5 & 1 & 1.5 & 2 & 2.5 & 3 & 3.5 & 4
    \\ \hline
    \multirow{3}{*}{820 RPM - load}
     & Precision & 0.174 & 0.236 & 0.289 & 0.323 & \bf{0.334} & 0.334 & 0.334 & 0.334 \\
     & Recall & 0.857 & 0.857 & 0.845 & 0.793 & \bf{0.758} & 0.695 & 0.641 & 0.596 \\
     & F1 & 0.289 & 0.371 & 0.431 & 0.459 & \bf{0.464} & 0.451 & 0.440 & 0.428 \\
    \hline
    \multirow{3}{*}{820 RPM - no load}
     & Precision & 0.157 & 0.213 & 0.259 & 0.285 & \bf{0.297} & 0.297 & 0.297 & 0.297 \\
     & Recall & 0.855 & 0.855 & 0.855 & 0.800 & \bf{0.770} & 0.706 & 0.652 & 0.605 \\
     & F1 & 0.265 & 0.341 & 0.397 & 0.421 & \bf{0.429} & 0.418 & 0.408 & 0.399 \\
    \hline
    \multirow{3}{*}{1500 RPM - load}
     & Precision & 0.145 & \bf{0.164} & 0.172 & 0.172 & 0.172 & 0.172 & 0.172 & 0.172 \\
     & Recall & 0.623 & \bf{0.523} & 0.422 & 0.337 & 0.307 & 0.281 & 0.259 & 0.241 \\
     & F1 & 0.235 & \bf{0.250} & 0.244 & 0.228 & 0.220 & 0.213 & 0.207 & 0.201 \\
    \hline
    \multirow{3}{*}{1500 RPM - no load}
     & Precision & 0.142 & 0.190 & 0.236 & 0.273 & \bf{0.283} & 0.283 & 0.283 & 0.283 \\
     & Recall & 0.855 & 0.855 & 0.855 & 0.832 & \bf{0.796} & 0.730 & 0.673 & 0.625 \\
     & F1 & 0.243 & 0.311 & 0.370 & 0.411 & \bf{0.418} & 0.408 & 0.399 & 0.390 \\
    \hline
    \multirow{3}{*}{Run up - load}
     & Precision & 0.245 & 0.325 & 0.390 & 0.444 & \bf{0.465} & 0.465 & 0.465 & 0.465 \\
     & Recall & 0.833 & 0.827 & 0.824 & 0.822 & \bf{0.815} & 0.747 & 0.689 & 0.640 \\
     & F1 & 0.378 & 0.467 & 0.530 & 0.577 & \bf{0.592} & 0.573 & 0.556 & 0.539 \\
    \hline
    \multirow{3}{*}{Run up - no load}
     & Precision & 0.232 & 0.302 & 0.361 & 0.410 & \bf{0.427} & 0.427 & 0.427 & 0.427 \\
     & Recall & 0.821 & 0.786 & 0.768 & 0.757 & \bf{0.737} & 0.676 & 0.624 & 0.579 \\
     & F1 & 0.361 & 0.436 & 0.491 & 0.532 & \bf{0.540} & 0.523 & 0.507 & 0.491 \\
    \hline
\end{tabular}

			\end{table}    
		
		\subsection{Variant 1: Time domain}
			\label{sec:uc1_var1}
			This variant evaluates the different operational conditions in the time domain representation. The goal is to detect any deviation on the current waveform with a generic comparison measure.
			
			\subsubsection{Block 1: Machine comparison}
				Due to the fact that the differences appearing in current amplitudes cannot be used in a fleet-based approach (\cref{fig:intermachine_difference}), min/max normalization is performed. Moreover, the current signals are downsampled to 50 samples per period. This removes the effect of signal noise and allows the analysis to focus on the waveforms.
				
\begin{figure}
					\centering
					\begin{subfigure}[t]{.5\textwidth}
						\centering
						\includegraphics[width=\linewidth]{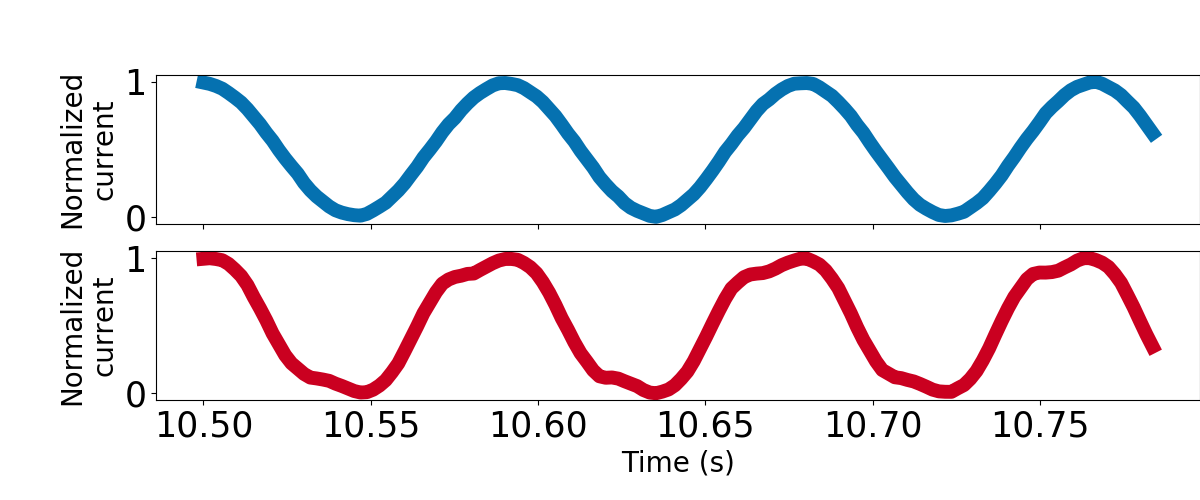}
						\caption{Measured current signals.}
						\label{fig:dtw_waveform_measured}
					\end{subfigure}%
					\begin{subfigure}[t]{.5\textwidth}
						\centering
						\includegraphics[width=\linewidth]{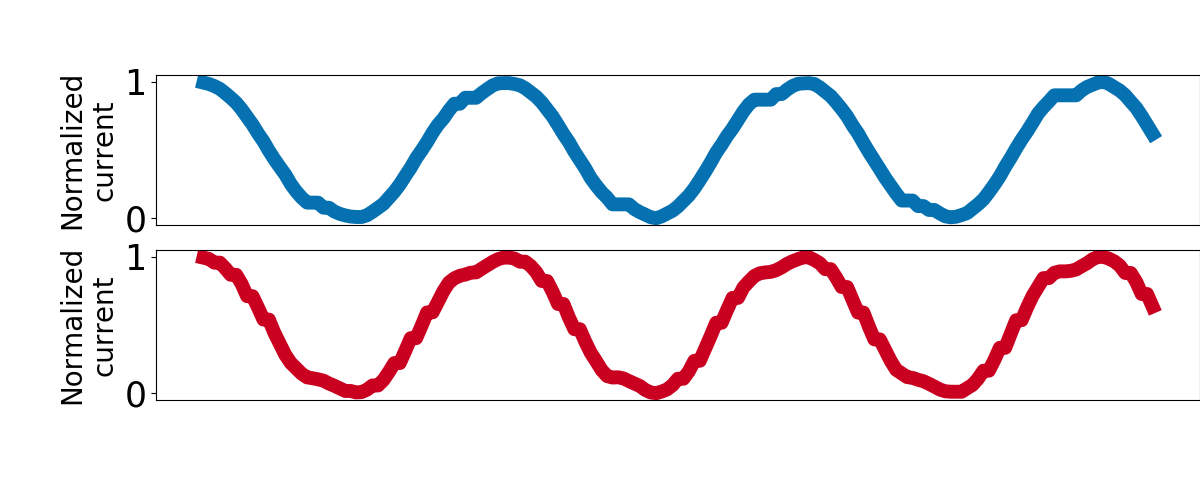}
						\caption{Warped current signals. The time axis is hidden, as the two signals have been non-linearly warped}
						\label{fig:dtw_waveform_warped}
					\end{subfigure}
					\caption{\ac{DTW} of a healthy (blue) and faulty (red) machine. While the measured current signals have a clear difference in shape (\subref{fig:dtw_waveform_measured}), \ac{DTW} can non-linearly align them so the signals become similar (\subref{fig:dtw_waveform_warped}).}
					\label{fig:dtw_waveform}
				\end{figure}

				The waveform of a machine affected by a voltage unbalance depends on the operational speed. A robust technique is thus required that detects any difference in this shape. \ac{DTW} is a popular technique for pattern recognition, but the classic implementation considers waveforms of healthy and faulty machines as similar (\cref{fig:dtw_waveform}). However, aligning healthy and faulty signals involves significant amounts of warping. In contrast, little warping occurs when comparing two healthy machines. Therefore, the amount of warping (\cref{eq:block1_dtwwarping}) can be used as a powerful indicator for waveform deviations, regardless of the exact shape. The $\Psi$-\ac{DTW} extension avoids unnecessary warping in the boundary points. Unaligned start- and endpoints would otherwise dominate the \ac{DTW} warping amount measure. The used value of $\Psi$ is corresponding to half a period of the current signal, allowing to align any phase offset.
				
			\subsubsection{Block 2: Fleet clustering}
				\label{sec:uc1_var1_block2}
				Hierarchical clustering, combined with a cophenetic correlation partition procedure, generates machine clusters. First,  hierarchical clustering uses pairwise machine comparisons and single linkage to construct a cluster hierarchy. Single linkage considers the minimum distance as the similarity of two subclusters. The cophenetic correlation procedure partitions this hierarchy with recursive top-down evaluations, if the cophenetic correlation is larger than $thr_{cc}$ (\cref{algo:partition_dendrogram}).
				
				The value of $thr_{cc}$ affects the effectiveness and the performance of the method but its optimal value is application dependent. Therefore, the framework is evaluated for various values of $thr_{cc}$ to show the parameter's impact on predictive performance. A lower value results in more clusters, which leads to more machines incorrectly labeled as faulty (low precision, higher false alarm rate). A higher value, on the other hand, leads to more misses of faulty machines (low recall, higher missed detection rate).
			
			\subsubsection{Block 3: Anomaly detection}
				\label{sec:uc1_var1_block3}
				The anomaly score is defined as the fraction of machines \changed{outside} each cluster. This is a simple scoring mechanism to demonstrate the concept of anomaly detection. At least \nicefrac{2}{3} of the machines are assumed to be healthy. The anomaly threshold $thr_{ad}$, \changed{above} which a machine is predicted as faulty, is thus set equal to \changed{\nicefrac{2}{3}}. If required, more advanced anomaly detection techniques can be used.
			
			\subsubsection{Block 4: Visualization}
				\label{sec:uc1_var1_block4}
				Visualizations allow a domain expert to interpret the framework's predictions. By showing the results of each block, an expert can confirm the predictions and avoid taking unneeded actions. All use-cases use the default implementation of \cref{sec:visualization}, being a (normalized) signal domain representation, machine dissimilarity, clustering, and anomaly scoring.
			
			\subsubsection{Results}
				\changed{The example of \cref{fig:results_template}} shows the evaluation of a single analysis window. A domain expert can use this visualization to validate the predictions or to tune the general framework. The time series show deviating behavior of machines D1\_2 and D2\_10 in each of the subfigures corresponding to the different building blocks.

				The evaluation procedure uses performance metrics to evaluate the framework's ability to detect a deviating waveform (\cref{tab:pu_curr_time}). Both the benchmark (\cref{tab:pu_curr_harmonics}) and this implementation are able to detect an unbalance, except at 1500 RPM in \changed{a} loaded condition due to the decrease of the flux. In other stationary scenarios, this implementation outperforms the benchmark method. Similar results are obtained in dynamic run-up scenarios. However, this implementation of the framework offers the advantage that it detects any deviations in the current waveforms, not only caused by an increase of the third harmonic.
		\clearpage		
				\begin{table}
					\caption {Performance of the proposed framework for fault detection using electrical signature analysis via waveform comparison. Scores corresponding to a scenario's highest F1-score are indicated in bold. \changed{For reference, the last column shows the performance of the classic diagnostics approach (\cref{tab:pu_curr_harmonics}) with the highest F1-score. In five out of the six cases, the proposed approach obtains the best F1 score.}}
					\label{tab:pu_curr_time}
					\changed{\begin{tabular}{ l  l r r r r r r |c  }
    & & \multicolumn{ 6 }{ c }{ $thr_{cc}$ } \\
    & Metric & 0.5 & 0.7 & 0.8 & 0.85 & 0.9 & 0.95
 & \thead{Classic diagnostics\\approach}    \\ \hline
    \multirow{3}{*}{820 RPM - load}
     & Precision & 0.218 & 0.542 & 0.696 & 0.932 & \bf{1.000} & 1.000     & \it{0.334} \\
     & Recall & 0.857 & 0.857 & 0.857 & 0.857 & \bf{0.857} & 0.812     & \it{0.758} \\
     & F1 & 0.348 & 0.664 & 0.768 & 0.893 & \bf{0.923} & 0.897     & \it{0.464} \\
    \hline
    \multirow{3}{*}{820 RPM - no load}
     & Precision & 0.189 & 0.573 & 0.922 & \bf{1.000} & \bf{1.000} & \bf{1.000}     & \it{0.297} \\
     & Recall & 0.855 & 0.855 & 0.855 & \bf{0.855} & \bf{0.855} & \bf{0.855}     & \it{0.770} \\
     & F1 & 0.309 & 0.686 & 0.887 & \bf{0.922} & \bf{0.922} & \bf{0.922}     & \it{0.429} \\
    \hline
    \multirow{3}{*}{1500 RPM - load}
     & Precision & \bf{0.135} & 0.043 & 0.000 & 1.000 & 1.000 & 1.000     & \it{0.164} \\
     & Recall & \bf{0.527} & 0.027 & 0.000 & 0.000 & 0.000 & 0.000     & \it{0.523} \\
     & F1 & \bf{0.215} & 0.034 & 0.000 & 0.000 & 0.000 & 0.000     & \it{0.250} \\
    \hline
    \multirow{3}{*}{1500 RPM - no load}
     & Precision & 0.382 & 0.959 & \bf{1.000} & \bf{1.000} & 1.000 & 1.000     & \it{0.283} \\
     & Recall & 0.855 & 0.855 & \bf{0.855} & \bf{0.855} & 0.818 & 0.245     & \it{0.796} \\
     & F1 & 0.528 & 0.904 & \bf{0.922} & \bf{0.922} & 0.900 & 0.394     & \it{0.418} \\
    \hline
    \multirow{3}{*}{Run up - load}
     & Precision & 0.479 & 0.697 & 0.821 & \bf{1.000} & \bf{1.000} & 1.000     & \it{0.465} \\
     & Recall & 0.852 & 0.852 & 0.852 & \bf{0.852} & \bf{0.852} & 0.778     & \it{0.815} \\
     & F1 & 0.613 & 0.767 & 0.836 & \bf{0.920} & \bf{0.920} & 0.875     & \it{0.592} \\
    \hline
    \multirow{3}{*}{Run up - no load}
     & Precision & 0.118 & 0.686 & \bf{1.000} & 1.000 & 1.000 & 1.000     & \it{0.427} \\
     & Recall & 0.857 & 0.857 & \bf{0.679} & 0.643 & 0.643 & 0.643     & \it{0.737} \\
     & F1 & 0.207 & 0.762 & \bf{0.809} & 0.783 & 0.783 & 0.783     & \it{0.540} \\
    \hline
\end{tabular}
}
				\end{table}

		\subsection{Variant 2: Frequency domain}
			This variant demonstrates the case in which detailed domain knowledge is included. A voltage unbalance affects the third harmonic of the fundamental frequency, which is used in this variant to compare the machines. In general, this is very similar to the benchmark method, as both depend on domain knowledge. 
			
			\subsubsection{Block 1: Machine comparison}
				A \ac{FFT} converts time series within an analysis window to their frequency data, which is converted to a log scale and normalized using min/max scaling. No outliers are present in the considered data set, other cases might need more robust techniques. Machines are represented in the feature space, by the amplitude or their current's third harmonic. This increases if one of the machines is faulty, as shown in \cref{sec:pu_curr_sota}.
				
				The periodic nature of current signals \changed{allow for estimating the fundamental frequency}. This corresponds to the highest peak in the frequency spectrum. The third harmonic is found at three times this frequency. The maximum value in a small window of $\pm$ 5Hz around the estimated frequency is used as its amplitude.
				
				This variant compares a machine pair by the difference in their third harmonic amplitudes (\cref{eq:block1_order}). 
				
				\begin{equation} \label{eq:block1_order}
				s(X, Y) = | harmonic(3, X) - harmonic(3, Y) |
				\end{equation}
			\subsubsection{Blocks 2 -- 4: Fleet clustering, Anomaly detection \& Visualization}
				Clustering, anomaly detection, and visualization make use of the default implementations. Hierarchical clustering combined with cophenetic correlation clusters the fleet. Therefore, the framework is evaluated for various values of $thr_{cc}$ to show the parameter's impact on predictive performance. Anomaly scores use these clusters and are defined as the fraction of machines \changed{outside} each cluster. The implementation considers a score \changed{above \nicefrac{2}{3}} as anomalous behavior. The visualization shows the frequency spectra enriched with the harmonic amplitudes windows.
			
			\subsubsection{Results}
\begin{figure}
					\centering
					\includegraphics[width=\textwidth]{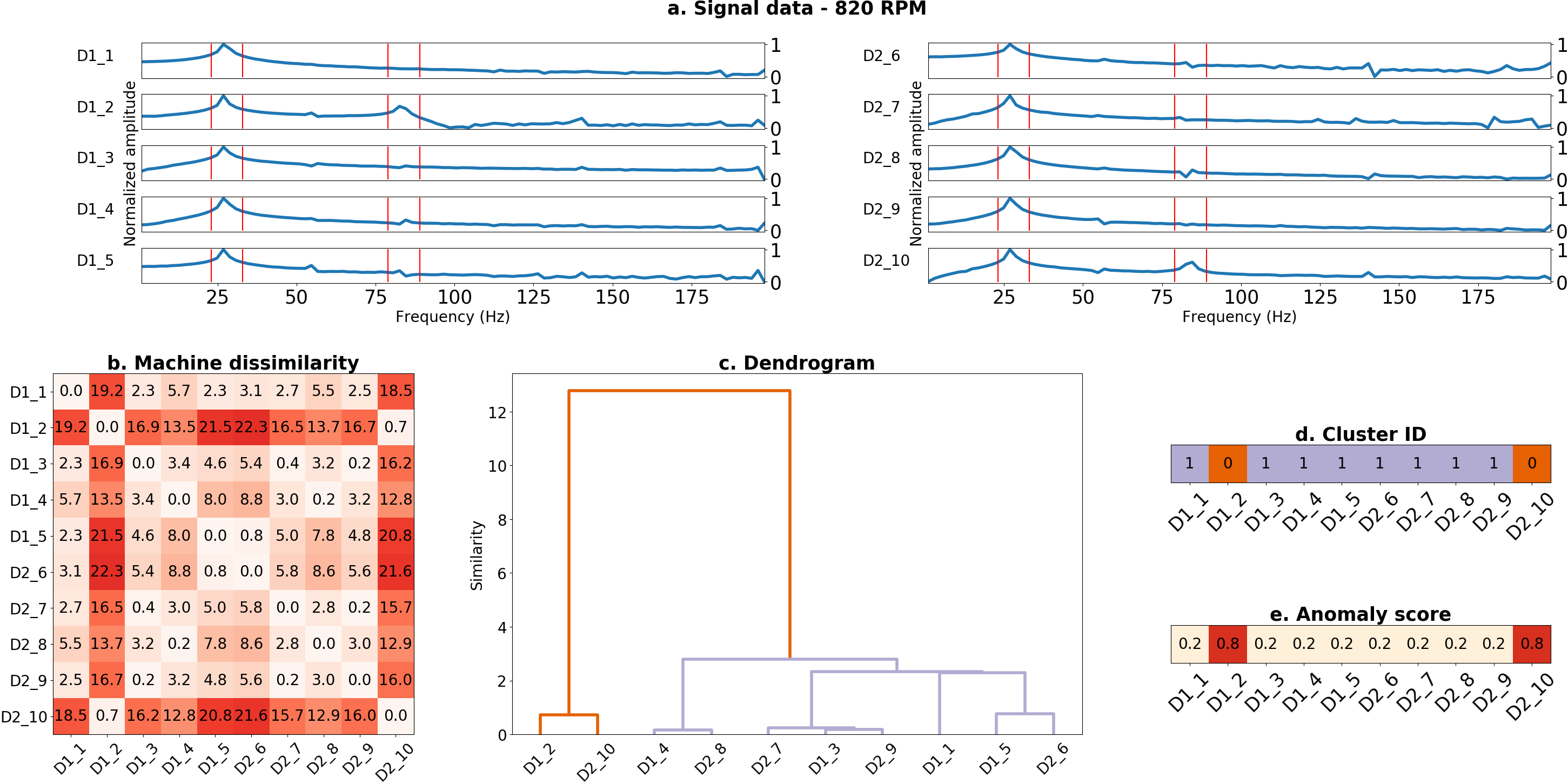}
					\caption{Visualization of frequency domain signals (a.), machine similarities (b.), clustering (c. \& d.) and anomaly scores (e.). The red vertical line in (a.) indicate the small windows of 10Hz around the fundamental and third harmonic frequencies. Its amplitudes are determined by each window's maximum amplitude. Clusters are partitioned with $thr_{cc} = 0.9$. Each machine is running stationary at 820 RPM. The framework correctly identifies D1\_2 and D2\_10 as anomalous.}
					\label{fig:result_pu_curr_freq_35}
				\end{figure}

				\Cref{fig:result_pu_curr_freq_35} shows the evaluation of a single analysis window. The anomalous behavior of D1\_2 and D2\_10 is successfully detected.
				
				The evaluation procedure uses performance metrics to evaluate the framework's ability to detect deviating third harmonic amplitudes (\cref{tab:pu_curr_freq}). This framework implementation has similar performance compared to the previous variant (\cref{tab:pu_curr_time}). It outperforms the benchmark method (\cref{tab:pu_curr_harmonics}) in almost all stationary operational conditions. At 1500 RPM in loaded condition, both methods show poor behavior. In the dynamic run-up scenarios\changed{,} on the other hand, the benchmark method is very competitive. It shows a higher recall (less missed detections), but its precision is slightly lower (more false alarms). This results in F1-scores in favor of the benchmark method. 
				
				\begin{table}
					\caption{Performance of the proposed framework for fault detection using electrical signature analysis via harmonic amplitudes. Scores corresponding to a scenario's highest F1-score are indicated in bold. \changed{For reference, the last column shows the performance of the classic diagnostics approach (\cref{tab:pu_curr_harmonics}) with the highest F1-score. In five out of the six cases, the proposed approach obtains the best F1 score.}}
					\label{tab:pu_curr_freq}
					\changed{\begin{tabular}{ l  l r r r r r r |c  }
    & & \multicolumn{ 6 }{ c }{ $thr_{cc}$ } \\
    & Metric & 0.5 & 0.7 & 0.8 & 0.85 & 0.9 & 0.95
 & \thead{Classic diagnostics\\approach}    \\ \hline
    \multirow{3}{*}{820 RPM - load}
     & Precision & 0.097 & 0.287 & 0.865 & \bf{1.000} & 1.000 & 1.000     & \it{0.334} \\
     & Recall & 0.857 & 0.857 & 0.857 & \bf{0.857} & 0.643 & 0.045     & \it{0.758} \\
     & F1 & 0.174 & 0.430 & 0.861 & \bf{0.923} & 0.783 & 0.085     & \it{0.464} \\
    \hline
    \multirow{3}{*}{820 RPM - no load}
     & Precision & 0.087 & 0.307 & 0.895 & \bf{1.000} & 1.000 & 1.000     & \it{0.297} \\
     & Recall & 0.855 & 0.855 & 0.855 & \bf{0.855} & 0.627 & 0.009     & \it{0.770} \\
     & F1 & 0.158 & 0.452 & 0.874 & \bf{0.922} & 0.771 & 0.018     & \it{0.429} \\
    \hline
    \multirow{3}{*}{1500 RPM - load}
     & Precision & \bf{0.098} & 0.118 & 0.250 & 1.000 & 1.000 & 1.000     & \it{0.164} \\
     & Recall & \bf{0.855} & 0.309 & 0.009 & 0.000 & 0.000 & 0.000     & \it{0.523} \\
     & F1 & \bf{0.176} & 0.171 & 0.018 & 0.000 & 0.000 & 0.000     & \it{0.250} \\
    \hline
    \multirow{3}{*}{1500 RPM - no load}
     & Precision & 0.097 & 0.250 & 0.712 & 0.887 & 0.979 & \bf{1.000}     & \it{0.283} \\
     & Recall & 0.855 & 0.855 & 0.855 & 0.855 & 0.855 & \bf{0.855}     & \it{0.796} \\
     & F1 & 0.175 & 0.387 & 0.777 & 0.870 & 0.913 & \bf{0.922}     & \it{0.418} \\
    \hline
    \multirow{3}{*}{Run up - load}
     & Precision & 0.100 & 0.535 & \bf{1.000} & \bf{1.000} & 1.000 & 1.000     & \it{0.465} \\
     & Recall & 0.852 & 0.852 & \bf{0.741} & \bf{0.741} & 0.444 & 0.259     & \it{0.815} \\
     & F1 & 0.179 & 0.657 & \bf{0.851} & \bf{0.851} & 0.615 & 0.412     & \it{0.592} \\
    \hline
    \multirow{3}{*}{Run up - no load}
     & Precision & 0.100 & 0.218 & \bf{1.000} & 1.000 & 1.000 & 1.000     & \it{0.427} \\
     & Recall & 0.857 & 0.857 & \bf{0.607} & 0.429 & 0.429 & 0.000     & \it{0.737} \\
     & F1 & 0.179 & 0.348 & \bf{0.756} & 0.600 & 0.600 & 0.000     & \it{0.540} \\
    \hline
\end{tabular}
}
				\end{table}

		\clearpage
		\subsection{Variant 3: \changed{T}ime/\changed{F}requency domain}
			This variant demonstrates a hybrid approach, in which a domain expert expects a fault to manifest itself in the frequency content, without specifying the exact frequency. It enables fault detection in a large range of working conditions. In this case, the framework is implemented to detect deviations in the spectrogram, a time/frequency domain representation.
			
			\subsubsection{Block 1: \changed{Machine} comparison}
				Each machine is represented by its spectrogram, a sequence of \ac{FFT}s obtained from consecutive windows. This implementation converts the data on a log scale and normalizes using min/max scaling \ac{FFT} to have a similar value ranges for each machine. A low-pass filter at 200 Hz removes unrelated high frequency content.
				
				\ac{DTW} (\cref{eq:block1_dtw}) is preferred as dissimilarity measure over Euclidean distance. If the drivetrain's speed fluctuates, \ac{DTW} will optimally align the spectrograms. In that case, these are considered as $n$-dimensional time series, with $n$ the number of frequency bins. Each of these bins is equally weighted in this implementation. If desired, a user could opt to have different weighting for frequency bins of interest.
				
				A high frequency resolution can result in false predictions due to the curse of dimensionality \cite{Friedman1997}. In that case, noise over unaffected frequencies would be larger than the amplitude increase in a single frequency bin. The task is however only to perform fault detection. Since the exact frequency at which a deviation occurs is not of interest, it is safe to have a low frequency resolution. In this variant, the selected \ac{FFT} window size is 0.05s. 
			
			\subsubsection{Blocks 2 -- 4: Fleet clustering, Anomaly detection \& Visualization}
				This variant reused the default clustering, anomaly detection, and visualization implementations. The first block uses hierarchical clustering and cophenetic correlation. Therefore, the framework is evaluated for various values of $thr_{cc}$ to show the parameter's impact on predictive performance. The second considers the fraction of machines \changed{outside} each cluster as anomaly scores. A score \changed{above \nicefrac{2}{3}} is considered as anomalous behavior. Finally, the visualization presents the signal data as a spectrogram.
			
			\subsubsection{Results}
\begin{figure}
					\centering
					\includegraphics[width=\textwidth]{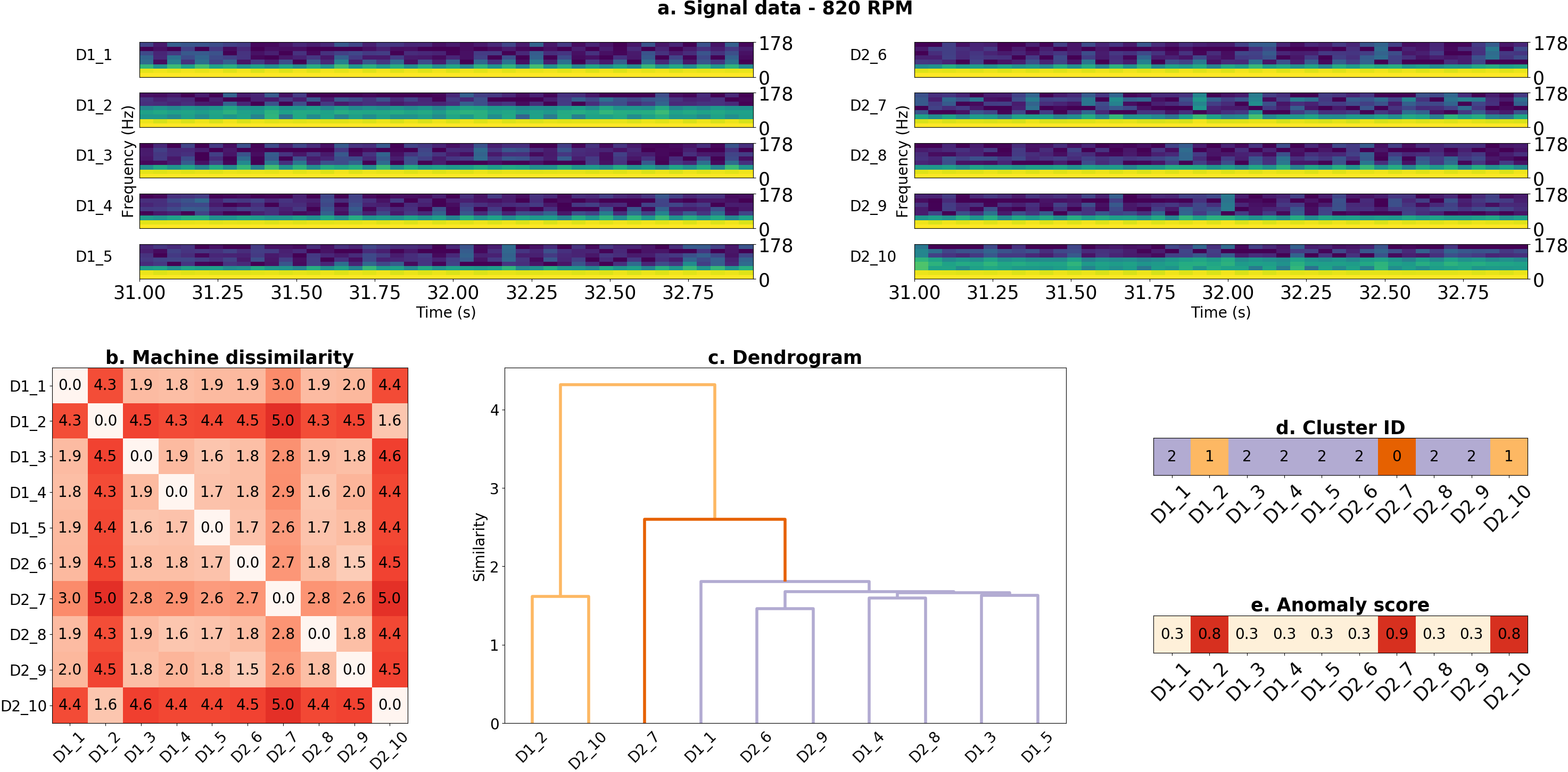}
					\caption{Visualization of time/frequency domain signals (a.), machine similarities (b.), clustering (c. \& d.) and anomaly scores (e.). The frequency resolution in (a.) is course, to avoid the curse of dimensionality while clustering. Clusters are partitioned with $thr_{cc} = 0.9$. Each machine is running stationary at 820 RPM.  The framework correctly identifies D1\_2 and D2\_10 as anomalous and incorrectly partitions D2\_7 by itself. This suggests that the choice of $thr_{cc}$ might not be optimal.}
					\label{fig:result_pu_curr_timefreq_35}
				\end{figure}

				\Cref{fig:result_pu_curr_timefreq_35} shows the evaluation of a single analysis window. A low-resolution spectrogram shows different machine behavior for faulty machines D1\_2 and D2\_10. Moreover, the dendrogram suggests using a lower value for $thr_{cc}$, as the healthy machine D2\_7 appears in its own cluster. 
				
				The evaluation procedure uses performance metrics to evaluate the framework's ability to detect deviating spectrograms (\cref{tab:pu_curr_timefreq}). It outperforms the benchmark method (\cref{tab:pu_curr_harmonics}) in most stationary scenarios. None of the methods detect an unbalance at 1500 RPM in \changed{a} loaded condition. In the dynamic run-up scenarios\changed{,} on the other hand, both methods have similar results. In general, this variant offers the advantage that it does not require domain knowledge or handcrafted rules.
				
				\begin{table}
					\caption{Performance of the proposed framework for fault detection using electrical signature analysis via spectrogram analysis. Scores corresponding to a scenario's highest F1-score are indicated in bold. \changed{For reference, the last column shows the performance for the classic diagnostics approach (\cref{tab:pu_curr_harmonics}) with the highest F1-score. In five out of the six cases, the proposed approach has the highest F1 score.}}
					\label{tab:pu_curr_timefreq}
					\changed{\begin{tabular}{ l  l r r r r r r |c  }
    & & \multicolumn{ 6 }{ c }{ $thr_{cc}$ } \\
    & Metric & 0.5 & 0.7 & 0.8 & 0.85 & 0.9 & 0.95
 & \thead{Classic diagnostics\\approach}    \\ \hline
    \multirow{3}{*}{820 RPM - load}
     & Precision & 0.157 & 0.505 & 0.738 & 0.906 & \bf{0.990} & 1.000     & \it{0.334} \\
     & Recall & 0.857 & 0.857 & 0.857 & 0.857 & \bf{0.857} & 0.768     & \it{0.758} \\
     & F1 & 0.265 & 0.636 & 0.793 & 0.881 & \bf{0.919} & 0.869     & \it{0.464} \\
    \hline
    \multirow{3}{*}{820 RPM - no load}
     & Precision & 0.236 & 0.913 & \bf{1.000} & \bf{1.000} & 1.000 & 1.000     & \it{0.297} \\
     & Recall & 0.855 & 0.855 & \bf{0.809} & \bf{0.809} & 0.618 & 0.091     & \it{0.770} \\
     & F1 & 0.370 & 0.883 & \bf{0.894} & \bf{0.894} & 0.764 & 0.167     & \it{0.429} \\
    \hline
    \multirow{3}{*}{1500 RPM - load}
     & Precision & \bf{0.106} & 0.198 & 0.000 & 1.000 & 1.000 & 1.000     & \it{0.164} \\
     & Recall & \bf{0.627} & 0.155 & 0.000 & 0.000 & 0.000 & 0.000     & \it{0.523} \\
     & F1 & \bf{0.182} & 0.173 & 0.000 & 0.000 & 0.000 & 0.000     & \it{0.250} \\
    \hline
    \multirow{3}{*}{1500 RPM - no load}
     & Precision & 0.208 & 0.839 & \bf{1.000} & 1.000 & 1.000 & 1.000     & \it{0.283} \\
     & Recall & 0.855 & 0.855 & \bf{0.855} & 0.636 & 0.118 & 0.000     & \it{0.796} \\
     & F1 & 0.335 & 0.847 & \bf{0.922} & 0.778 & 0.211 & 0.000     & \it{0.418} \\
    \hline
    \multirow{3}{*}{Run up - load}
     & Precision & 0.178 & 0.487 & \bf{0.864} & 1.000 & 1.000 & 1.000     & \it{0.465} \\
     & Recall & 0.852 & 0.704 & \bf{0.704} & 0.630 & 0.593 & 0.333     & \it{0.815} \\
     & F1 & 0.295 & 0.576 & \bf{0.776} & 0.773 & 0.744 & 0.500     & \it{0.592} \\
    \hline
    \multirow{3}{*}{Run up - no load}
     & Precision & 0.221 & 0.900 & 0.900 & \bf{1.000} & 1.000 & 1.000     & \it{0.427} \\
     & Recall & 0.679 & 0.643 & 0.643 & \bf{0.643} & 0.393 & 0.000     & \it{0.737} \\
     & F1 & 0.333 & 0.750 & 0.750 & \bf{0.783} & 0.564 & 0.000     & \it{0.540} \\
    \hline
\end{tabular}
}
				\end{table}                
	\clearpage
	\section{Use-case 2: Voltage unbalance -- Vibration signatures}
		\label{sec:use_case_vu_acc}
		This use-case involves vibration measurements of an electrical machine fleet, similar to the previous use-case. The task is to detect a voltage unbalance fault based on accelerometer signals. Accelerometers are often already present for condition monitoring of other (mechanical) components, hence it is valuable if they can also be used for voltage unbalance detection. This section illustrates that it is indeed feasible and moreover shows how the framework can use multiple sensor channels and indicators of interest.
		
		\subsection{Experimental fleet test rig and data set}
\begin{figure}
				\centering
				\begin{subfigure}[t]{.6\textwidth}
					\centering
					\includegraphics[width=\linewidth]{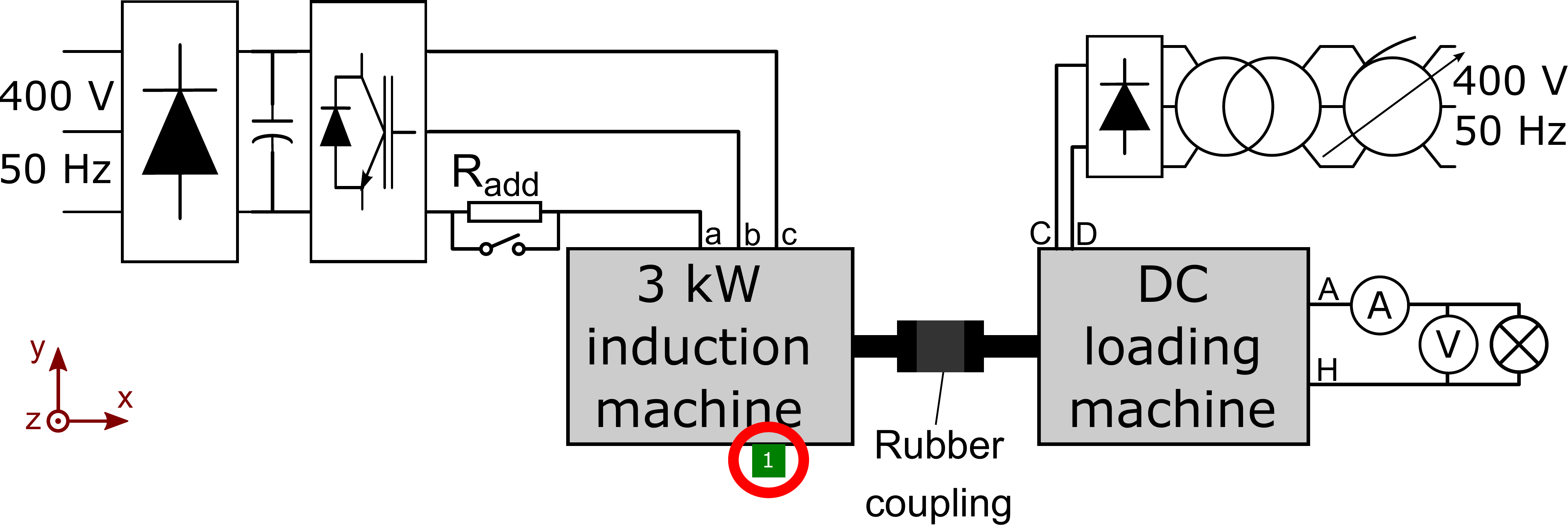}
					\caption{Illustration of drivetrain D1\_2.}
					\label{fig:ULB_bench_acc}
				\end{subfigure}%
				\begin{subfigure}[t]{.4\textwidth}
					\centering
					\includegraphics[width=0.9\linewidth]{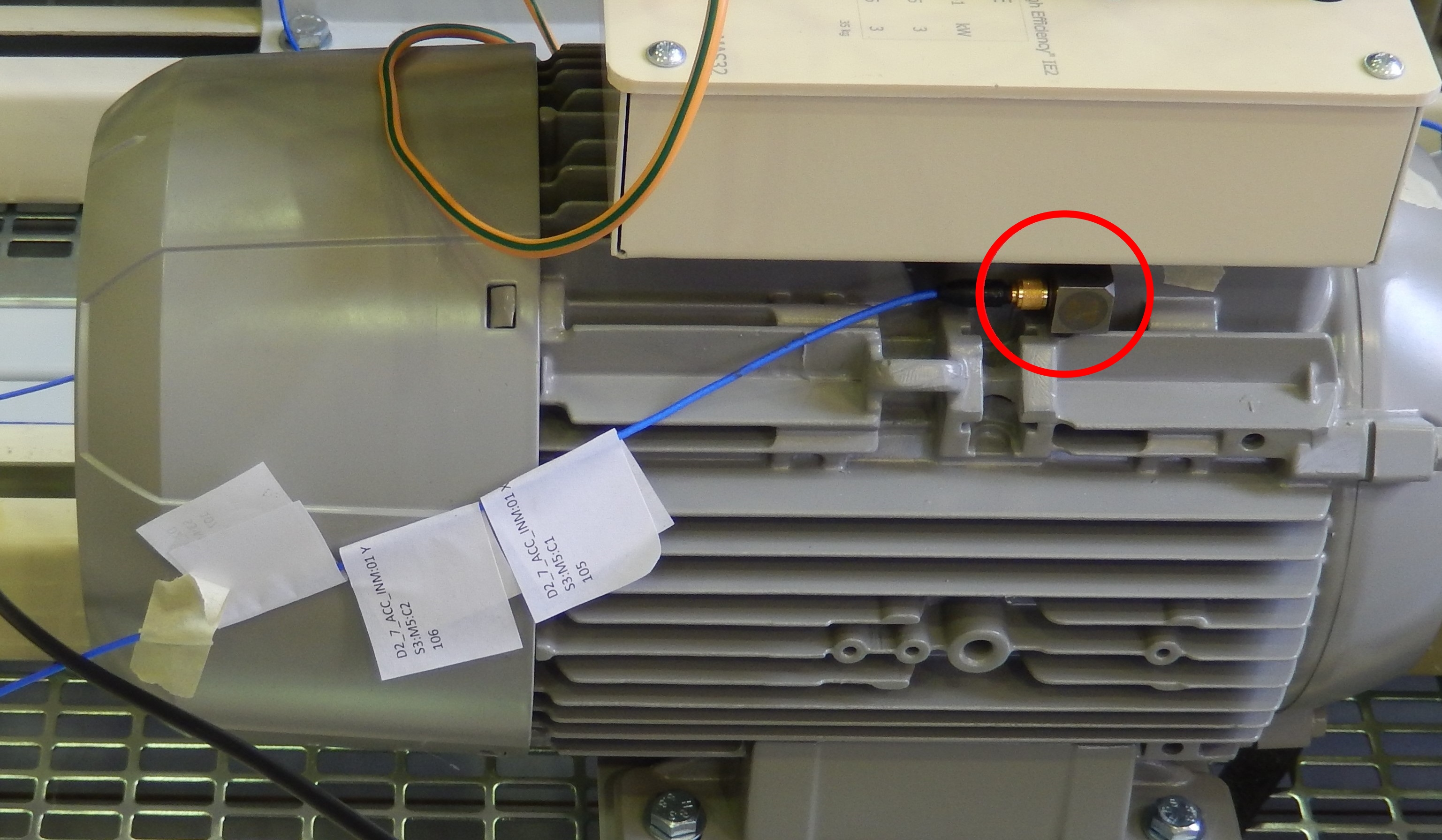}
					\caption{Drive side motor.}
					\label{fig:ULB_bench_acc_sensor}
				\end{subfigure}
				\caption{A 3D accelerometer (indicated in red) is mounted on the drive side motor and is sampled at 12800Hz.}
				\label{fig:ULB_bench_acc_setup}
			\end{figure}

			This use-case considers a setup of ten electrical drivetrains. A voltage unbalance is again emulated with an external resistor. PCB ICP AC accelerometers measure vibrations. One sensor is mounted on the side of the drive side motor (\cref{fig:ULB_bench_acc_setup}) and is sampled at 12800 Hz. Operational speed is measured with a single tachometer mounted on D1\_2, triggered once every rotation. This is representative for the full fleet, as all drivetrains are set up to always have the same speed, controlled by an ABB drive controller. One exception is machine D2\_8, whose run-up parameters are set incorrectly. For this reason, this drivetrain is not used in the evaluation procedures when considering a run-up.
			
		\subsection{Classic voltage unbalance diagnosis techniques}
			\label{sec:pu_acc_sota}
			For a healthy, symmetrical, motor, electrodynamic forces have no alternating components. In the balanced case, the forces are rotating with the flux. In \changed{the} case of unbalance they have a component rotating in the opposite direction. This results in a pulsation of the amplitude of the forces at twice the stator frequency. This causes radial vibrations in the induction motor \cite{Mollet2018,Maruthi2006,Tavner2008}. The forces have a tangential component (creating the torque) but also a radial component (the stator flux attracts the rotor like an electromagnet) generating the vibrations.
			
\begin{figure}
				\centering
				\begin{subfigure}[t]{.5\textwidth}
					\centering
					\includegraphics[width=\linewidth]{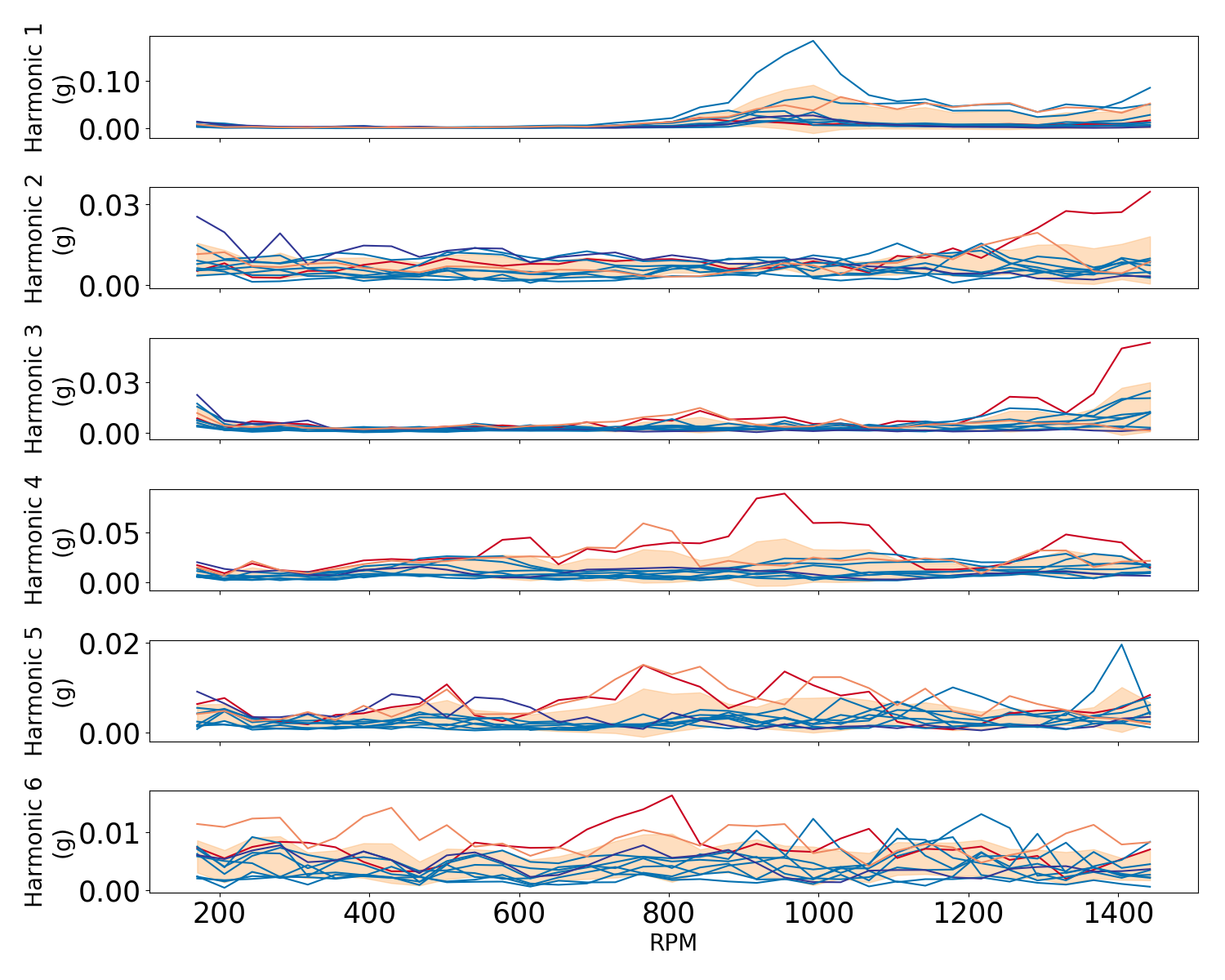}
					\caption{Without load}
					\label{fig:pu_acc_orders_overview_noload}
				\end{subfigure}%
				\begin{subfigure}[t]{.5\textwidth}
					\centering
					\includegraphics[width=\linewidth]{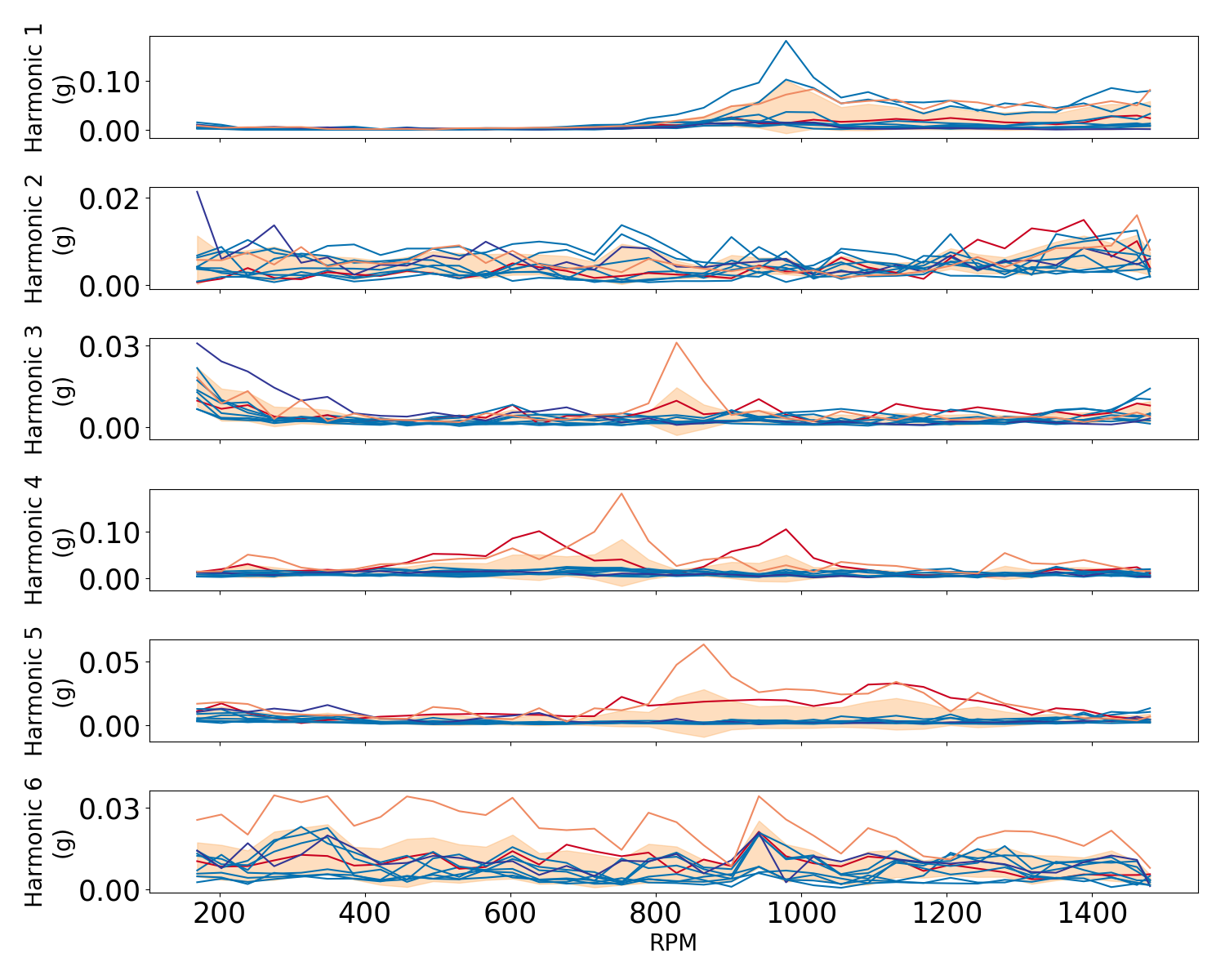}
					\caption{With load}
					\label{fig:pu_acc_orders_overview_load}
				\end{subfigure}
				\caption{Harmonics 1 (top) -- 6 (bottom) of the X-axis accelerometer in a run-up. The faulty drivetrains are shown in blue (D1\_2) and red (D2\_10). The optimal indicator to detect the voltage unbalance depends on the machine of interest and the operational speed. The orange band indicates the range in which a machine is considered healthy, being 2 standard deviations around the mean.}
				\label{fig:pu_acc_orders_overview}
			\end{figure}

			Detecting voltage unbalance based on vibration signatures is an example of a condition monitoring case that requires different thresholds, as operational conditions affect vibration amplitudes. For example, additional load or operating at a resonance frequency leads to an increase, while lower operating speeds result in lower vibration levels (\cref{fig:pu_acc_orders_overview}). Moreover, many frequencies and sensor directions can be considered. Detailed machine knowledge is required to know these frequencies a priori.
			
			The machine response varies for D1\_2 and D2\_10, the faulty drive trains (\cref{fig:pu_acc_orders_overview}). Their resonance frequencies differ and in loaded conditions, D2\_10 is significantly more affected by the fault compared to D1\_2. The optimal indicator depends thus not only on operational conditions but also on the evaluated machine.
			
			The remainder of this use-case utilizes a simple condition monitoring method as a benchmark to validate the proposed framework. This method considers a machine as healthy when the amplitude of all its harmonics 3 -- 6 is within $\sigma$ standard deviations of the fleet's mean. If one harmonic is outside this range, the machine is predicted as being faulty. Note that this is a simplified fleet approach, which avoids setting threshold manually. \Cref{tab:pu_acc_harmonics} shows the fault detection performance for this methodology with different values of $\sigma$, \Cref{fig:pu_acc_orders_overview} illustrates the specific case for $\sigma = 2$. 
			
			\begin{table}
				\caption{Performance of classic voltage unbalance detection using accelerometer harmonics 3 -- 6. Scores corresponding to a scenario's highest F1-score are indicated in bold.}
				\label{tab:pu_acc_harmonics}
				\begin{tabular}{ l  l r r r r r r r r  }
    & & \multicolumn{ 8 }{ c }{ $\sigma$ } \\
    & Metric & 0.5 & 1 & 1.5 & 2 & 2.5 & 3 & 3.5 & 4
    \\ \hline
    \multirow{3}{*}{820 RPM - load}
     & Precision & 0.110 & 0.143 & 0.168 & 0.191 & \bf{0.201} & 0.201 & 0.201 & 0.201 \\
     & Recall & 0.871 & 0.871 & 0.817 & 0.784 & \bf{0.772} & 0.708 & 0.653 & 0.607 \\
     & F1 & 0.195 & 0.246 & 0.279 & 0.307 & \bf{0.319} & 0.314 & 0.308 & 0.302 \\
    \hline
    \multirow{3}{*}{820 RPM - no load}
     & Precision & 0.104 & 0.116 & 0.125 & \bf{0.132} & 0.133 & 0.133 & 0.133 & 0.133 \\
     & Recall & 0.871 & 0.753 & 0.648 & \bf{0.569} & 0.526 & 0.483 & 0.445 & 0.414 \\
     & F1 & 0.185 & 0.201 & 0.210 & \bf{0.214} & 0.212 & 0.208 & 0.205 & 0.201 \\
    \hline
    \multirow{3}{*}{1500 RPM - load}
     & Precision & 0.098 & 0.101 & \bf{0.107} & 0.107 & 0.106 & 0.106 & 0.106 & 0.106 \\
     & Recall & 0.847 & 0.730 & \bf{0.623} & 0.510 & 0.463 & 0.425 & 0.392 & 0.364 \\
     & F1 & 0.176 & 0.177 & \bf{0.183} & 0.176 & 0.173 & 0.170 & 0.167 & 0.165 \\
    \hline
    \multirow{3}{*}{1500 RPM - no load}
     & Precision & 0.098 & 0.107 & 0.115 & \bf{0.119} & 0.119 & 0.119 & 0.119 & 0.119 \\
     & Recall & 0.871 & 0.837 & 0.736 & \bf{0.620} & 0.564 & 0.517 & 0.477 & 0.443 \\
     & F1 & 0.176 & 0.189 & 0.199 & \bf{0.199} & 0.196 & 0.193 & 0.190 & 0.187 \\
    \hline
    \multirow{3}{*}{Run up - load}
     & Precision & 0.246 & 0.301 & \bf{0.349} & 0.367 & 0.368 & 0.368 & 0.368 & 0.368 \\
     & Recall & 0.817 & 0.756 & \bf{0.705} & 0.611 & 0.558 & 0.512 & 0.473 & 0.439 \\
     & F1 & 0.378 & 0.431 & \bf{0.467} & 0.458 & 0.444 & 0.428 & 0.414 & 0.400 \\
    \hline
    \multirow{3}{*}{Run up - no load}
     & Precision & 0.249 & 0.295 & \bf{0.322} & 0.332 & 0.332 & 0.332 & 0.332 & 0.332 \\
     & Recall & 0.848 & 0.789 & \bf{0.670} & 0.562 & 0.511 & 0.469 & 0.433 & 0.402 \\
     & F1 & 0.385 & 0.429 & \bf{0.434} & 0.418 & 0.403 & 0.389 & 0.376 & 0.364 \\
    \hline
\end{tabular}

			\end{table}

		\subsection{Variant 1: Frequency domain}
			This variant demonstrates the use of multiple sensors. The implementation combines each of the accelerometer directions (X, Y, Z) in the comparison measure. Moreover, it considers domain knowledge by multiple harmonic frequencies for each of these directions. A data-driven approach automatically selects the best indicator value.
		
		\subsubsection{Block 1: Machine comparison}
			Machines are represented by harmonic frequencies 3 -- 6 of each sensor direction. The implementation uses the tachometer to define the fundamental frequency. A \ac{FFT} converts the time domain measurements to frequency spectra, in which the maximum value within a window of $\pm$5Hz around the estimated frequencies is considered as their amplitude. Together, these form a sequence with length 1 and 12 dimensions (3 sensor directions, 4 harmonics).
			
			Accelerometer signals can be noisy and require a robust normalization procedure. This variant uses percentile scaling, as data can be affected by outliers. In this case, normalization also makes sure to equally weight every harmonic. Otherwise, a harmonic with a larger range could dominate the clustering. 
			
			The Euclidean distance measures the differences between a machine pair (\cref{eq:block1_euclidean_distance}).
		
		\subsubsection{Blocks 2 -- 4: \changed{Fleet} clustering, anomaly detection \& visualization}
			This variant reuses the default clustering, anomaly detection, and visualization blocks. The first combines hierarchical clustering with a cophenetic correlation procedure. Therefore, the framework is evaluated for various values of $thr_{cc}$ to show the parameter's impact on predictive performance. The second uses the fraction of machines in each cluster as the anomaly score. A score \changed{above \nicefrac{2}{3}} is considered as anomalous behavior. Finally, the frequency spectra of each drivetrain are shown combined with the considered harmonic windows.
		
		\subsubsection{Results}
			\Cref{fig:result_pu_curr_freq_35} shows the evaluation of a single analysis window. The anomalous behavior of D1\_2 and D2\_10 is successfully detected in each block.
			
			The evaluation procedure uses performance metrics to evaluate the framework's ability to detect faults based on multiple sensor directions and harmonics (\cref{tab:pu_acc_drive_freq}). In loaded scenarios, it outperforms the benchmark method (\cref{tab:pu_acc_harmonics}). Both approaches show poor results in unloaded scenarios, which can be explained by the reduced vibrations in the absence of a load. At 1500 RPM, none of the methods detect the fault due to a decrease of the flux. Lower general performance (compared to electrical signature analysis, \cref{sec:use_case_vu_curr}) in dynamic run-up scenarios is caused by the fault not manifesting itself in the vibration signature at every operational speed. In general, this implementation offers the advantage that it can consider multiple harmonics, without needing individual threshold values. This is an advantage especially when the optimal indicator varies with the operational conditions.
			
			\begin{table}
				\caption{Performance of the proposed framework for fault detection using vibration signature analysis via harmonic amplitudes. \changed{For reference, the last column shows the performance for the classic diagnostics approach (\cref{tab:pu_acc_harmonics}) with the highest F1-score. In five out of the six cases, the proposed approach has the highest F1 score.}}
				\label{tab:pu_acc_drive_freq}
				\changed{\begin{tabular}{ l  l r r r r r r |c  }
    & & \multicolumn{ 6 }{ c }{ $thr_{cc}$ } \\
    & Metric & 0.5 & 0.7 & 0.8 & 0.85 & 0.9 & 0.95
 & \thead{Classic diagnostics\\approach}    \\ \hline
    \multirow{3}{*}{820 RPM - load}
     & Precision & 0.121 & 0.462 & 0.545 & 0.565 & 0.610 & \bf{0.908}     & \it{0.201} \\
     & Recall & 0.871 & 0.871 & 0.871 & 0.871 & 0.871 & \bf{0.871}     & \it{0.772} \\
     & F1 & 0.213 & 0.603 & 0.671 & 0.686 & 0.718 & \bf{0.889}     & \it{0.319} \\
    \hline
    \multirow{3}{*}{820 RPM - no load}
     & Precision & 0.097 & 0.153 & 0.185 & 0.196 & \bf{0.205} & 0.163     & \it{0.132} \\
     & Recall & 0.629 & 0.452 & 0.435 & 0.435 & \bf{0.435} & 0.266     & \it{0.569} \\
     & F1 & 0.168 & 0.229 & 0.260 & 0.270 & \bf{0.278} & 0.202     & \it{0.214} \\
    \hline
    \multirow{3}{*}{1500 RPM - load}
     & Precision & \bf{0.091} & 0.004 & 0.000 & 0.000 & 0.000 & 0.000     & \it{0.107} \\
     & Recall & \bf{0.508} & 0.008 & 0.000 & 0.000 & 0.000 & 0.000     & \it{0.623} \\
     & F1 & \bf{0.154} & 0.005 & 0.000 & 0.000 & 0.000 & 0.000     & \it{0.183} \\
    \hline
    \multirow{3}{*}{1500 RPM - no load}
     & Precision & 0.105 & 0.133 & \bf{0.178} & 0.172 & 0.116 & 0.000     & \it{0.119} \\
     & Recall & 0.702 & 0.435 & \bf{0.379} & 0.258 & 0.129 & 0.000     & \it{0.620} \\
     & F1 & 0.183 & 0.204 & \bf{0.242} & 0.206 & 0.122 & 0.000     & \it{0.199} \\
    \hline
    \multirow{3}{*}{Run up - load}
     & Precision & 0.345 & 0.710 & \bf{1.000} & 1.000 & 1.000 & 1.000     & \it{0.349} \\
     & Recall & 0.857 & 0.786 & \bf{0.696} & 0.536 & 0.536 & 0.429     & \it{0.705} \\
     & F1 & 0.492 & 0.746 & \bf{0.821} & 0.698 & 0.698 & 0.600     & \it{0.467} \\
    \hline
    \multirow{3}{*}{Run up - no load}
     & Precision & 0.240 & \bf{0.508} & 0.469 & 0.469 & 0.200 & 0.143     & \it{0.322} \\
     & Recall & 0.857 & \bf{0.589} & 0.411 & 0.411 & 0.107 & 0.071     & \it{0.670} \\
     & F1 & 0.375 & \bf{0.545} & 0.438 & 0.438 & 0.140 & 0.095     & \it{0.434} \\
    \hline
\end{tabular}
}
			\end{table}
	
	\clearpage
	\section{Conclusion}
		\label{sec:conclusion}
		Machine fleet condition monitoring offers several advantages over both traditional condition monitoring approaches and supervised machine learning techniques. First, it removes the need to know all potential machine faults a priori, as the fleet-based approach can detect any deviation in machine behavior. However, the framework can incorporate domain knowledge if available.
		Second, fleet monitoring does not require a high-quality historical (labeled) data set for training. Both traditional signal processing approaches as supervised machine learning need this data to learn faulty machine behavior. Moreover, these require ground-truth knowledge about the machine's actual health status. This is not required for fleet monitoring, as it assumes the majority of the machines to be healthy.
		Third, fleet monitoring allows analysis in dynamic operational conditions. Online comparisons allow detecting deviating machine behavior even in unconsidered operational conditions.
		Finally, the framework offers a high level of interpretability. Visualizations allow a domain expert to get insights into the predictions and gain confidence in the methodology. This is especially in contrast with black box machine learning techniques, whose models are very hard to interpret.

	\section*{Acknowledgments}
		The authors acknowledge the financial support of VLAIO (Flemish Innovation \& Entrepreneurship) through the Baekeland PhD mandate [nr. HBC.2017.0226]; the O\&O project REFLEXION [nr. IWT. 150334]. Jesse Davis is partially supported by the KU Leuven research funds [C14/17/070] and Research Foundation - Flanders [EOS No. 30992574].  Jesse Davis, Wannes Meert and Konstantinos Gryllias receive funding from the Flemish Government under the "Onderzoeksprogramma Artificiële Intelligentie (AI) Vlaanderen" programme.
	
	\bibliographystyle{unsrt}
	\bibliography{main}
\end{document}